\def\paperID{10386}
\def\confName{ICCV}
\def\confYear{2025}

\def\paperTitle{Improving Large Vision and Language Models by Learning \\ from a Panel of Peers}

\def\authorBlock{
    Jefferson Hernandez$^{1}$\thanks{Work done while interning at Adobe Research.}, Jing Shi$^{2}$, Simon Jenni$^{2}$, Vicente Ordonez$^{1}$, Kushal Kafle$^{2}$  \\
    $^{1}$Rice University, $^{2}$Adobe Research\\
    {\tt\small jefehern@rice.edu, \{jingshi, jenni, kkafle\}@adobe.com, vicenteor@rice.edu} 
}

\newif\ifreview 
\newif\ifarxiv \newcommand{\arxiv}{\arxivtrue}
\newif\ifcamera 
\newif\ifrebuttal 

\newcommand{\bon}{Bo$N$\xspace}
\newcommand{\pop}{PoP\xspace}

\newcommand{\icon}[2][2.2ex]{% default 1.2ex is perfect for captions
  \raisebox{-0.5\height}{\includegraphics[height=#1]{#2}}%
}
\newcommand{\MIS}[1][2.2ex]{\icon[#1]{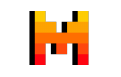}}
\newcommand{\VIC}[1][2.2ex]{\icon[#1]{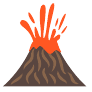}}
\newcommand{\LLA}[1][2.2ex]{\icon[#1]{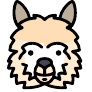}}
\arxiv % \review OR \arxiv OR \cameraready

\pdfoutput=1
\documentclass[10pt,twocolumn,letterpaper]{article}
\ifreview \usepackage[review]{iccv} \fi
\ifarxiv \usepackage[pagenumbers]{iccv} \fi
\ifrebuttal \usepackage[rebuttal]{iccv} \fi
\ifcamera \usepackage{iccv} \fi

\usepackage{graphicx}	
\usepackage{amsmath}	
\usepackage{amssymb}	
\usepackage{booktabs}
\usepackage{times}
\usepackage{microtype}
\usepackage{epsfig}
\usepackage{pifont}
\usepackage[table,xcdraw,dvipsnames]{xcolor}
\usepackage{caption}
\usepackage{float}
\usepackage{placeins}
\usepackage{color, colortbl}
\usepackage{stfloats}
\usepackage{enumitem}
\usepackage{tabularx}
\usepackage{xstring}
\usepackage{multirow}
\usepackage{xspace}
\usepackage{url}
\usepackage{subcaption}
\usepackage{xcolor}
\usepackage[hang,flushmargin]{footmisc}
\usepackage{afterpage}
\usepackage{float}

\ifcamera \usepackage[accsupp]{axessibility} \fi

\usepackage{tcolorbox}
\tcbuselibrary{skins}
\newtcolorbox{prompt}[1]{
    enhanced,
    left=4mm,
    right=4mm,
    top=2mm,
    bottom=2mm,
    boxsep=0mm,
    rounded corners,
    title=#1,
    fontupper=\footnotesize\linespread{0.9}\fontfamily{lmr}\selectfont,
    }

% ...

\newcommand{\cmark}{{\ding{51}}}%
\newcommand{\xmark}{{\ding{55}}}%

\newcommand{\supp}{supplemental material\xspace}
\ifarxiv \renewcommand{\supp}{appendix\xspace} \fi

\newcommand{\R}[1]{{%
    \textbf{%
        \ifstrequal{#1}{1}{\textcolor{red}{5yXu}}{%
        \ifstrequal{#1}{2}{\textcolor{blue}{bfE4}}{%
        \ifstrequal{#1}{3}{\textcolor{magenta}{sTPX}}{%
        \ifstrequal{#1}{4}{\textcolor{teal}{R#1}}{%
                           \textcolor{cyan}{R#1}%
        }}}}%
    }%
}}
\usepackage{xr-hyper}

\makeatletter
\newcommand*{\addFileDependency}[1]{
  \typeout{(#1)}
  \@addtofilelist{#1}
  \IfFileExists{#1}{}{\typeout{No file #1.}}
}

\makeatother

\definecolor{iccvblue}{rgb}{0.21,0.49,0.74}
\usepackage[pagebackref,breaklinks,colorlinks,citecolor=iccvblue]{hyperref}
\usepackage[capitalize]{cleveref}
\crefname{section}{Sec.}{Secs.}
\crefname{table}{Table}{Tables}
\crefname{figure}{Fig.}{Figs.}
\begin{document}
%% TITLE
\title{\paperTitle}
\author{\authorBlock}
\maketitle

\begin{abstract}
Traditional alignment methods for Large Vision and Language Models (LVLMs) primarily rely on human-curated preference data. Human-generated preference data is costly; machine-generated preference data is limited in quality; and self-supervised preference data often introduces hallucinations. To overcome these limitations, we propose a novel Panel-of-Peers learning framework inspired by collaborative learning among humans. This approach leverages a panel of LVLMs, each evaluating and learning from their collective outputs through an iterative self-improvement process. By simulating a peer review system, our models generate, assess, and refine outputs in response to a curated set of prompts, mimicking a classroom learning environment. We demonstrate that this methodology enhances model performance without requiring extensive human-labeled datasets. Our experiments show significant improvement across multiple benchmarks, demonstrating the potential of peer evaluations as a scalable alternative to self-supervised alignment. Notably, we show that Panel-of-Peers increases the average score on fifteen benchmarks from 48\% to 57\%.
\end{abstract}

\section{Introduction}
\label{sec:intro}
Large Vision and Language Models (LVLMs) have demonstrated impressive capabilities that require a diverse set of skills such as compositional reasoning (e.g.~\cite{liu2024visual, liu2023improvedllava}), use of general knowledge (e.g.~\cite{team2023gemini, achiam2023gpt}), pictorial reasoning about abstract figures (e.g.~\cite{radford2021learning}), and character recognition (e.g.~\cite{yuan2021florence}). Learning a generalist model that can tackle all these tasks at once has been a challenge, as different LVLMs might feature complementary strengths depending on the richness of their training datasets. A successful approach for improving both Large Language Models (LLMs) and LVLMs has been to further refine them by relying on human preference data to ensure that their outputs are more aligned with the expectations of human users~\cite{rafailov2024direct, ouyang2022training}.

\begin{figure}[t!]
    \centering
    \includegraphics[width=0.9\linewidth]{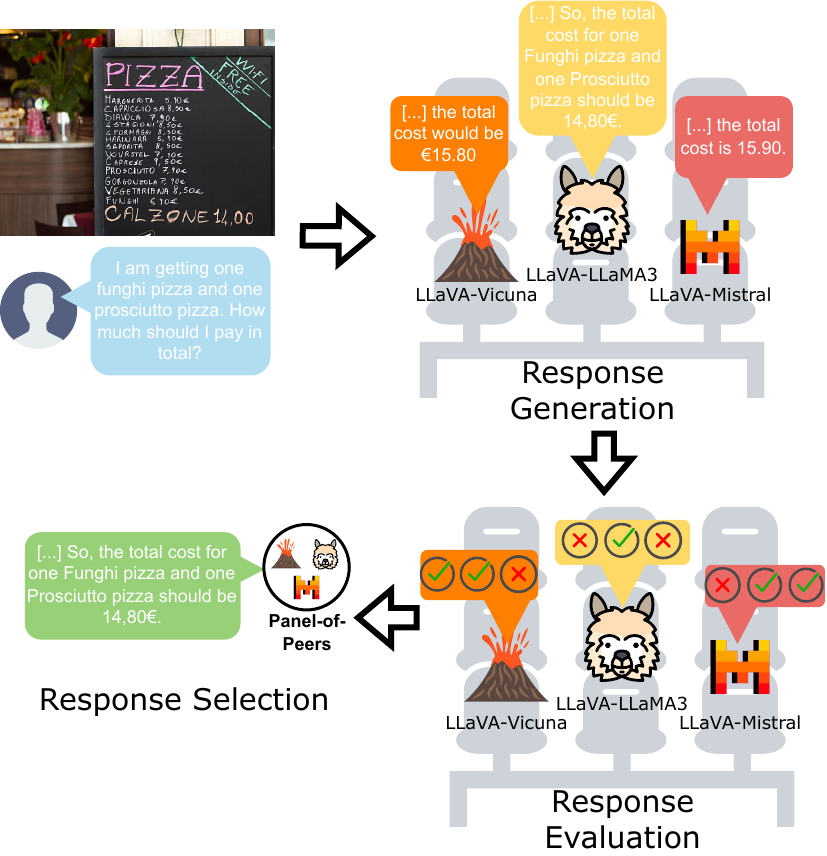}
    \caption{A \textbf{Panel-of-Peers (\pop)} generates candidate responses from multiple LVLMs. The panel's scoring of these responses is used to build a preference set, which is used to tune one or all the members of the panel, improving their accuracy individually. \pop significantly outperforms other forms of reaching consensus across many benchmarks.}
    \label{fig:panel}
    \vspace{-0.2in}
\end{figure}

The training process for LVLMs has largely converged into three stages: 1) unimodal pretraining, where the vision encoder~\cite{radford2021learning, zhai2023sigmoid, oquab2023dinov2} and the LLM~\cite{llama3model, jiang2023mistral, vicuna2023} are independently trained on a large corpus of data; 2) multimodal pretraining, where the unimodal models are combined and trained on a large corpus of image-text data~\cite{schuhmann2022laion, ordonez2011im2text}, sometimes augmented with extra knowledge (\eg, OCR, or grounding data); and 3) a supervised fine-tuning (SFT)~\cite{liu2023improvedllava, liu2024llavanext} stage, where the model is trained on domain-specific data to enhance its performance on various downstream tasks. Despite extensive research into this three-stage pipeline, LVLMs continue to face challenges, including hallucination issues, misalignment between visual and textual representations, and persistent knowledge gaps. These limitations underscore the need to further refine our approach to enhance the alignment and reliability of LVLMs.

Collecting high-quality multimodal data can be one of the most straightforward solutions but doing so on a large scale is often expensive, and therefore most methods use low-quality, large-scale captioned image-text pairs, followed by fine-tuning with small-scale, higher-quality, supervised data. Many recent studies rely on learning from machine-generated data from foundation models (\eg~GPT-4V~\cite{achiam2023gpt}, Gemini~\cite{team2023gemini}). Still, the performance of the resulting models is limited by the performance of the foundation models~\cite{zhao2023svit, wang2023see} and fails to significantly reduce the cost~\cite{wu2024gpt, wang2023see}.

\begin{figure*}[t!]
    \centering
    \includegraphics[width=0.95\textwidth]{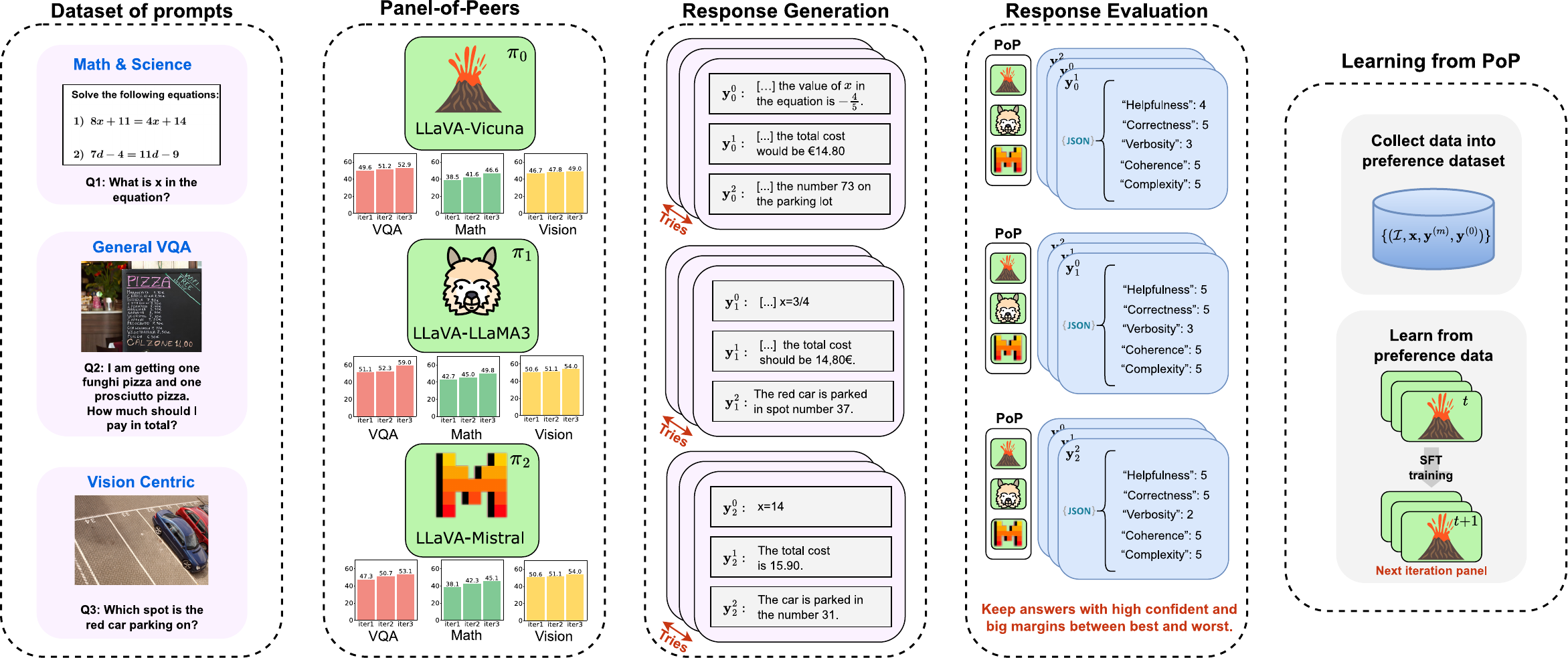}
    \caption{\textbf{Illustration of the overall learning from feedback from peers approach}. Our post-alignment strategy involves rejection sampling, supervised finetuning, and preference optimization methods to learn from peers. See text for details.}
    \label{fig:model}
\vspace{-0.01in}
\end{figure*}

In this work, we describe a method for LVLMs to continue learning from peer-to-peer feedback, which we denote Panel-of-Peers (\pop) that uses a list of tasks/questions with no answers. This method takes inspiration from how students learn in a classroom environment~\cite{o2014cognitive}. They are taught the basics in a chapter and are asked to complete a series of exercises that pose challenging questions without answers. By going through the exercises and \emph{discussing with their peers}, they further solidify their understanding and emerge with an improved ability to complete the related tasks. 

Our peer-to-peer training methodology is summarized in Figure~\ref{fig:model}. 
We begin by constructing a panel of peers (i.e., models of roughly the same capacity, trained on the same training set), but take advantage of the fact that some models might be naturally better for some tasks than their peers. Then, each model in the panel produces candidate answers for a dataset of new prompts (without ground truth answers) and also evaluates the correctness of each other's answers. These evaluations are then combined into a final reward score using reward ensembling methods~\cite{costereward}. The resulting preference data is then used to fine-tune all the models in the panel (See Figure~\ref{fig:panel}). This process results in the same number of models as in the original panel, all of which can be used for individual inference. Our extensive experiments show that such feedback from peers can augment the performance of the entire panel and individual LVLM members within the panel. Additionally, it can teach an individual panel member a previously unseen task, provided that the other members possess that knowledge. % \\

\noindent
\textbf{Our main contributions are as follows:}
\begin{itemize}
    \item We introduce Learning from a Panel-of-Peers (\pop), a novel self-improvement paradigm for bootstrapping the capabilities of a set of peer LVLMs.
    \item We show that \pop increases the average score on 15 selected benchmarks from 48 to 57 (9 absolute points).
    \item We demonstrate that \pop can enable knowledge transfer from panel members possessing certain abilities (e.g., OCR) to members that do not have that ability.
    \item We conduct extensive ablations regarding the choice of reward ensembling methods, the size/ability of panel members, allowing each member to produce one or multiple candidate answers, and the choice of alignment objectives to clearly show the effects of our modeling choices.
\end{itemize}

\section{Related Work}
\label{sec:related}
Our work is related to self-improving methods in LLMs, preference alignment in LVLMs, and the use of models as judges for the output of other models.

\noindent{\bf LVLMs-as-a-judge.} 
Strong LVLMs like GPT-4V~\cite{achiam2023gpt} have been widely used to evaluate vision-language tasks~\cite{liu2023mmbench, yu2024mmvet} through both pointwise~\cite{liu2024visual, sun2023aligning} and paired evaluations~\cite{lu2024wildvision, yu2024rlaifv}. Typically, this involves crafting a ``scoring prompt'' to guide the evaluator in scoring, often referencing a \textit{golden} answer. Our work diverges by scoring answers solely based on the model's internal knowledge, foregoing reference answers. This aligns with efforts in \textit{golden-answer}-free evaluation, such as Self-Taught Evaluators~\cite{wang2024self} and others~\cite{yuan2024self, xu2024perfect}, which teach scoring through dedicated datasets. Additionally, the use of a panel of models for evaluation has shown promise. For example, PoLL~\cite{verga2024replacing} demonstrated that a panel of weaker models can produce human-aligned scores comparable to stronger models. Research suggests models excel at evaluation over generation~\cite{cobbe2021training, xu2024perfect, fang2024vila}, leveraging verification to self-improve. Prometheus-Vision~\cite{lee2024prometheusvision} and LLaVA-Critic~\cite{xiong2024llava} further advanced evaluation by training models with curated datasets or user-defined scoring criteria. In contrast, we show that with robust ``evaluation prompts,'' models require no additional training to perform evaluations. Furthermore, we unify answer generation and evaluation within the same model, updating it through iterative self-improvement rather than keeping evaluation models fixed.

\noindent{\bf Self-Improvement in Large Language Models.}
Various recent works have investigated self-improvement or self-play strategies. Self-Rewarding Language Models~\cite{yuan2024self} enable an LLM to act as both a reward model and an answer generator, creating and ranking self-instruction data. Self-Improving Robust Preference Optimization~\cite{choi2024self} refines an LLM's answers through in-context learning with a loss function akin to direct preference optimization~\cite{rafailov2024direct}. Self-play Fine-Tuning~\cite{chen2024self} frames self-improvement as a two-player game: one player generates answers indistinguishable from human-annotated data, while the other attempts to identify machine-generated answers. Self-Play Preference Optimization~\cite{wu2024self} similarly applies a two-player game where the LLM interacts with its prior version, using an exponential weight update to solve the Nash equilibrium and ensure convergence. Unlike these LLM-focused methods, our Panel-of-Peers approach directly addresses LVLM modality misalignment by integrating peer feedback with supervised fine-tuning for more effective alignment.

\noindent{\bf Alignment in Large Vision Language models.} 
LLMs require alignment to ensure behavior matches human preferences~\cite{ouyang2022training, ziegler2019fine, rafailov2024direct}. In LVLMs, alignment methods focus on reducing hallucinations and enhancing modality alignment. For instance, LLaVA-RLFH~\cite{sun2023aligning} collected a dataset of $\sim$10K human interactions to reduce hallucinations using RLFH on the \texttt{LLaVA-1.0-7b} model. POVID~\cite{zhou2024aligning} and SeVa~\cite{zhu2024self} simulate hallucinations by injecting errors, then aligning via DPO. 
Li~\etal~\cite{li2024multi} collects 5k image-question pairs and obtain four responses per pair, which are scored by a foundation model~\cite{team2023gemini} using the prompt from~\cite{wang2024helpsteer2}; alignment is performed via DPO.
SIMA~\cite{wang2024enhancing} creates two answers using different decoding algorithms and uses the model's own self-critic capabilities to create preference data for preference optimization.
STIC~\cite{deng2024enhancing} combines augmentations and adverse prompts to induce dispreferred data for DPO-based alignment, while CSR~\cite{zhou2024calibrated} uses beam search and CLIPScore to align responses by ranking preferred answers with DPO loss. VILA$^2$~\cite{fang2024vila} takes a task-specialized approach, constructing versions of the original model tailored to captioning, OCR, and general knowledge. Although these methods enhance alignment, they are resource-intensive, rely on human annotations, and may introduce biases. Notably, CSR~\cite{zhou2024calibrated} and VILA$^2$~\cite{fang2024vila} are closest to our work: CSR uses calibrated rewards based on answer-image CLIPScore, which may suffice for captions but is limited in tasks like OCR and VQA. In contrast, PoP leverages knowledge gaps within peer models to achieve alignment across tasks. Unlike VILA$^2$’s \textit{multiple teachers-single student} approaches, which require teacher re-training, PoP employs a \textit{multiple students-to-multiple students} setup, where models iteratively learn from each other’s knowledge gaps within the panel.

\section{Learning with a Panel-of-Peers}
\label{sec:method}
We focus on autoregressive LVLMs where image tokens are projected into the embedding space of textual information and concatenated, a method popularized by LLaVA~\cite{liu2024visual, liu2023improvedllava}. We start with models that have undergone the standard three-stage training (pretrain-align-SFT) in all our experiments. We begin by constructing a panel of models with similar performance (which we call peers) and devise a novel learning algorithm where all members of the panel self-improve their capabilities in a self-bootstrapping loop. Our method is called Panel-of-Peers (PoP). An overview of the stages of \pop is shown in Figure~\ref{fig:model}. 

Our method consists of two stages (1) candidate response generation and (2) data creation and fine-tuning, which take place alternately and can be used over multiple iterations. In the candidate response generation stage, we prompt the panel of peers with the image and question from a selected dataset designed for the models to learn a diverse set of tasks (captioning, OCR, general knowledge, among others). In the preference data creation and fine-tuning stage, each model performs peer-to-peer evaluations of the responses of the other models in the panel. A reward score is created by combining the evaluations from all the models in the panel, which is then used alongside rejection sampling to create a preference dataset. The final stage uses these data to fine-tune each model in the panel using a preference alignment algorithm; we perform preference fine-tuning on the synthetic data, which has been found to work better than supervised fine-tuning techniques. This process is repeated iteratively; each time, the panel generates, evaluates, and learns from its answers.

\subsection{Reward Modeling}

Inspired by consensus methods for multiple LLM judges~\cite{verga2024replacing, xu2024perfect} and other works in machine learning that show the benefit of model ensembles, we propose a peer-to-peer evaluation approach, where an ensemble of models scores each other's output. When acting as a judge, the model $\pi_{i}$ scores the output $\mathbf{y}_j$, \ie \emph{The car is parked in spot 31}, from one of its peer models $\pi_{j} (\mathbf{y}_j \vert \mathbf{x})$. The peer model is tasked with answering query $\mathbf{x}$, \ie \emph{Where is the car parked?} We create a prompt $\mathbf{p}$ that evaluates answers along five axes: Helpfulness, Correctness, Verbosity, Coherence, and Complexity~\cite{wang2024helpsteer, wang2024helpsteer2}, to create pseudo-rewards for learning. Each score is graded on a Likert scale from 1 to 5, summed, and then divided by 25 to normalize it to the 0-1 range, resulting in the final reward score. Following~\cite{verga2024replacing}, we experiment with single- and relative-point scoring, where the judge is tasked with rating the quality of a single model output independently or in comparison with other outputs. The judge receives natural language instructions on how the grading should be performed, detailed in full in the \supp, to create pseudo-rewards for learning. The rating is based solely on the judge's internal knowledge of what constitutes a quality output.  We construct a panel of $M$ peers. Each model independently assigns a reward value between 0 and 1 to a given peer output, resulting in a reward $R_i(\mathbf{y}_j)  = \pi_{i} (\cdot \vert \mathbf{p} \cup \mathbf{x} \cup \mathbf{y}_j)$, where $\mathbf{y}$ is the answer from the model, $\mathbf{x}$ is the original question, and $\mathbf{p}$ is the evaluation prompt. To combine individual reward scores from the panel, we use \emph{mean voting} $R_{\mu}(\mathbf{y}_j)$ to average the scores from all peers.

\subsection{Candidate Response Generation}
Similar to how we use a panel of peers to generate rewards and evaluate the output of an answer, our objective is to use the same panel to generate responses to build preference data. Each model in the panel receives the same input image and query pair and produces a candidate response. Then, for each response $\mathbf{y}$, we calculate its reward score $R_{\mu}(\mathbf{y})$. We apply an additional rejection sampling step before constructing the preference dataset, retaining only \textit{chosen} samples with a quality reward score of at least 0.85.
\noindent
We can also augment the responses generated using Best-of-$N$ sampling, which is an inference-only alignment algorithm that works as follows. Let $Y_N = \{\mathbf{y}^{(n)}\}^N_{n=1}$ be the multi-set containing $N$ i.i.d. samples from $\pi_{i} (\mathbf{y}_i \vert \mathbf{x})$ for a query $\mathbf{x}$.
 Then, \bon algorithm returns $\mathbf{y}^{\star}$, where
\begin{equation}
    \mathbf{y}^{\star} = \underset{\mathbf{y}^{(n)} \in Y_N}{\mathrm{argmax}} \mbox{  } R\left(\mathbf{y}^{(n)}\right).
\end{equation}
This method has been shown to be win-rate optimal and KL optimal asymptotically~\cite{beirami2024theoretical} to the preference alignment problem. Finally, after sampling $N$ answers using sentence-level beam search from each of the $M$ models in the panel, we end up with $N \times M$ candidate responses. 

\noindent
These two components, (1) candidate response generation and (2) reward modeling, constitute the \pop algorithm. We also explore using a single sample from each model, and this formulation is denoted as \textit{st-\pop}. 

\subsection{Preference Curation and Iterative Training}

After generating candidate responses with their reward scores, our next step is to curate a preference dataset. Here, for each input prompt, we select the responses with the highest and lowest cumulative reward scores as the preferred and dispreferred responses, respectively, to construct the preference dataset for fine-tuning. For each iteration $t$, we denote the constructed preference data as $\mathcal{D}_t = \{(\mathcal{I}, \mathbf{x}, \mathbf{y}^{(n)}, \mathbf{y}^{(1)})\}_{i=1}^{|\mathcal{D}|}$, where $\mathcal{I}$ represents the image, $\mathbf{x}$ the text prompt, and $\mathbf{y}^{(n)}, \mathbf{y}^{(1)}$ are the highest and lowest ranked answers by the panel of peers after filtering. After obtaining the preference data, we fine-tune the whole panel of Large Vision-Language Models (LVLM) using SimPO. We choose SimPO~\cite{meng2024simpo} because it employs an implicit reward formulation that directly aligns with the generation metric, eliminating the need for a reference model. Additionally, it introduces a target reward margin $\gamma$ to help separate the winning and losing responses.
\begin{equation}
    \begin{aligned}
     & \mathcal{L}_{\text{SimPO-\pop}}(\pi_{\theta_{t+1}}, \pi_{\theta_{t}}) = \\ &  -\mathbb{E}_{\mathcal{D}_t}  \left[ \log \sigma \left(\frac{\beta}{|\mathbf{y}^{(n)}|} \pi_{\theta_t}(\mathbf{y}^{(n)}) - \frac{\beta}{|\mathbf{y}^{(1)}|} \pi_{\theta_t}(\mathbf{y}^{(1)}) - \gamma \right)\right]
    \end{aligned}
\end{equation}
Where $\pi_{\theta_{t+1}}$ and $\pi_{\theta_{t}}$ represent the models from the next and previous iterations, respectively. This optimization is repeated for each member of the panel. For the next iteration, the entire process of (1) candidate response generation and (2) data creation and fine-tuning are repeated, with the panel always initialized from the checkpoints of the previous iteration. To evaluate the effect of the alignment objective on performance, we compare SimPO with Supervised Fine-Tuning (SFT) using the same curated preference dataset. 
\section{Experiment Settings}
\label{sec:experiments}
\noindent{\bf Implementation Details.} We use the original \texttt{LLaVA-1.5} recipe of two stages: (1) multimodal pre-training and (2) supervised fine-tuning. Unless otherwise specified, rewards and responses are generated using the same visual backbone and the following LLMs: Mistral-7B~\cite{jiang2023mistral}, Llama3-8B~\cite{llama3model}, and Vicuna-7B~\cite{vicuna2023}, resulting in three models for the panel of peers. These are trained on the \texttt{BLIP-LAION-CC-SBU-558k} dataset~\cite{liu2024visual} for stage 1, and the open source version of the \texttt{LLaVA-Instruct-mix665k} dataset~\cite{liu2023improvedllava} for stage 2. The images and prompts used to construct the data are randomly sampled from the \texttt{Cambrian-7M} dataset~\cite{tong2024cambrian}, keeping the original proportions of Language: 21.00\%, General Knowledge: 34.52\%, OCR: 27.22\%, Counting: 8.71\%, Math: 7.20\%, Code: 0.87\%, and Scientific Knowledge: 0.88\%.

We sample 15 responses from each member of the panel (\pop) and compare against sampling a single answer (st-\pop) as a baseline. We reject samples with a reward of less than 0.85 and maintain a margin of 0.75 between preferred and dispreferred answers, as we found that this helps preference optimization. This process creates a dataset with a total of $\sim$300K samples per self-improvement iteration out of a starting random sample of 1M images from the \texttt{Cambrian-7M}. It is worth noting that in each iteration, we start with more than the specified number of samples to ensure the required number is obtained after rejection sampling. Additionally, each iteration might include different samples. Overall, the iterative training is conducted over three iterations, using full fine-tuning on 8 A100 80GB GPUs. It takes roughly 80 hours to collect the preference data and 10 hours per model for the self-improvement step. For more detailed information on training hyperparameters and training data, please refer to the \supp.

\noindent{\bf Evaluation Benchmarks.}  We conducted evaluations using \texttt{VLMEvalKit}\footnote{\url{https://github.com/open-compass/VLMEvalKit}. Commit \texttt{f547007}}~\cite{duan2024vlmevalkit}. Specifically, we select the following datasets split into the following categories: \emph{Chart\&OCR}: ChartQA~\cite{masry2022chartqa}, OCR-Bench~\cite{liu2023hidden}, OCRVQA~\cite{mishra2019ocr}, TextVQA~\cite{Singh_2019_CVPR}; \emph{General VQA}: MMBench~\cite{liu2023mmbench},  MM-Vet~\cite{yu2024mmvet},  SEED-Bench~\cite{li2024seed}; \emph{Knowlegde}: AI2D~\cite{kembhavi2016diagram}, MMMU(val)~\cite{yue2024mmmu}, MMStar~\cite{chen2024we}, MathVista(val)~\cite{lu2024mathvista}, ScienceQA$^{\text{IMG}}$~\cite{lu2022learn}; \emph{Hallucination}: HallucinationBench~\cite{guan2023hallusionbench},  POPE~\cite{li2023evaluating}; \emph{Vision Centric}:  RealWorldQA~\cite{realworldqa2024}. More detailed descriptions of each dataset are discussed in the \supp.

\noindent{\bf Baselines.} We compare Panel-of-Peers Learning with the following preference learning approaches Silkie~\cite{li2023silkie}, LLaVA-RLHF~\cite{sun2023aligning}, RLHF-V~\cite{yu2024rlhf}, POVID~\cite{zhou2024aligning}, Self-Rewarding~\cite{zhou2024calibrated}, CSR~\cite{zhou2024calibrated}, SeVa~\cite{zhu2024self}, STIC~\cite{deng2024enhancing}, SIMA~\cite{wang2024enhancing}, and LLaVA-Critic~\cite{xiong2024llava} as well as, state-of-the-art methods taken from the OpenVLM Leaderboard (see \supp).\footnote{\url{https://huggingface.co/spaces/opencompass/open_vlm_leaderboard}. Accessed Oct 30, 2024.}.

\newcommand{\grayfont}{\color{gray}}

\begin{table}[t!]
\centering
\footnotesize
\setlength\tabcolsep{1.5pt}
\renewcommand{\arraystretch}{1.2}
\begin{tabular}{@{}l c c c c c c@{}}
\toprule
\textbf{Model} & \textbf{Data} & \textbf{MMB} & \textbf{SEED-B} & \textbf{MM-Vet} & \textbf{SQA} & \textbf{POPE} \\ 
\midrule
LLaVA-1.5-7B~\cite{liu2023improvedllava}      & -    & 64.3 & 58.6 & 30.5 & 66.8 & 85.9 \\ 
\mbox{ }+VLFeedback~\cite{li2023silkie}       & 80k  & 64.0 & 59.3 & 31.2 & 66.2 & 83.7 \\
\mbox{ }+RLHF~\cite{sun2023aligning}            & 10k  & 63.4 & 58.1 & 31.1 & 65.8 & 81.5 \\
\mbox{ }+RLHF-V~\cite{yu2024rlhf}              & 1.4k & 63.6 & 60.1 & 30.9 & 67.1 & 86.2 \\
\mbox{ }+POVID~\cite{zhou2024aligning}          & 17k  & 64.9 & 60.2 & 31.8 & 68.8 & \underline{86.9} \\
\mbox{ }+Self-Reward~\cite{zhou2024calibrated}   & 17k  & 64.5 & 60.0 & 31.4 & 69.6 & \underline{86.9} \\
\mbox{ }+CSR~\cite{zhou2024calibrated}           & 17k  & 65.4 & 60.3 & 33.9 & \underline{70.7} & \textbf{87.0} \\
\mbox{ }+SeVa~\cite{zhu2024self}                & 8k   & \underline{65.6} & 65.8 & \textbf{37.2} & 67.5 & 86.7 \\
\mbox{ }+STIC~\cite{deng2024enhancing}          & 6k   & 65.3 & \underline{66.2} & 32.6 & 67.4 & 85.8 \\ 
\mbox{ }+SIMA~\cite{wang2024enhancing}          & 17k  & 64.9 & 60.6 & 31.6 & 69.1 & - \\ 
\mbox{ }+LLaVA-Critic~\cite{xiong2024llava}      & 113k & 64.1 & 60.0 & 32.2 & 68.4 & 85.8 \\ 
\midrule
\rowcolor{gray!20} \multicolumn{7}{l}{\textit{Preference Optimization (iteration 1)}} \\ 
\mbox{ }+st-\pop-iter1~(ours) & 300k & 65.5 & 61.6 & 31.9 & 67.1 & 86.8 \\
\mbox{ }+\pop-iter1~(ours)                        & 300k & \textbf{68.7} & \textbf{67.9} & \underline{34.1} & \textbf{71.2} & \textbf{87.0} \\
\midrule
\rowcolor{gray!20} \multicolumn{7}{l}{\textit{Preference Optimization (iteration 3)}} \\ 
\mbox{ }+st-\pop-iter3~(ours) & 900k & 67.4 & 67.9 & 32.5 & 74.0 & 86.3 \\
\mbox{ }+\pop-iter3~(ours)                        & 900k & \textbf{72.5} & \textbf{68.8} & \underline{35.0} & \textbf{86.4} & \textbf{87.0} \\
\bottomrule
\end{tabular}
\caption{\textbf{Performance of \pop using {LLaVA-1.5}}. The evaluation benchmarks span several comprehensive evaluations, as well as general knowledge and hallucination evaluations. 
We mark the best performance in \textbf{bold}, and the second-best is \underline{underlined}.}
\label{tab:comparission}
\vspace{-0.2in}
\end{table}

\vspace{-0.1in}
\section{Results and Ablations}
\label{sec:results}

\definecolor{light-gray}{gray}{0.5}
\definecolor{light-green}{HTML}{D5F5E3}
\definecolor{light-blue}{HTML}{D6EAF8}
\newcommand{\padcol}{\multicolumn{1}{c}{}}
\begin{table*}[t!]
  \centering
  \small
  \setlength{\tabcolsep}{4pt}
  \renewcommand{\arraystretch}{1.2}
  \begin{tabular}{%
      ll                               % Capability, Benchmark
      ccc@{\hspace{6pt}}c@{\hspace{6pt}}% Iter-0  + spacer-1
      ccc@{\hspace{6pt}}c@{\hspace{6pt}}% Iter-1  + spacer-2
      ccc@{\hspace{6pt}}c@{\hspace{6pt}}% Iter-2  + spacer-3
      ccc}                             % Iter-3
  \toprule
  \multirow{2}{*}{\textbf{Capability}} &
  \multirow{2}{*}{\textbf{Benchmark}} &
  \multicolumn{3}{c}{\textbf{Iteration 0}} & \padcol &
  \multicolumn{3}{c}{\textbf{Iteration 1}} & \padcol &
  \multicolumn{3}{c}{\textbf{Iteration 2}} & \padcol &
  \multicolumn{3}{c}{\textbf{Iteration 3}}\\
  \cmidrule{3-5}\cmidrule{7-9}\cmidrule{11-13}\cmidrule{15-17}
        & & \MIS[3ex] & \VIC[3ex] & \LLA[3ex] & & % spacer-1
          \MIS[3ex] & \VIC[3ex] & \LLA[3ex] & & % spacer-2
          \MIS[3ex] & \VIC[3ex] & \LLA[3ex] & & % spacer-3
          \MIS[3ex] & \VIC[3ex] & \LLA[3ex] \\

% ─── sample row (showing the extra & for spacers) ─────────────
  \midrule
% General VQA
    \multirow{3}{*}{GeneralVQA} & MMBench~\cite{liu2023mmbench} 
    & 62.4 & 66.5 & 65.6 && 70.1 & 68.7 & 73.8 && 68.2 & 71.5 & 74.1 && 69.1 & 72.5 & 75.1 \\
    & MM-Vet~\cite{yu2024mmvet}
    & 21.1 & 32.9 & 26.2 && 26.6 & 34.1 & 32.9 && 31.1 & 32.6 & 33.0 && 32.3 & 35.0 & 35.4 \\
    & SEED-Bench~\cite{li2024seed} 
    & 64.6 & 65.8 & 61.6 && 69.2 & 67.9 & 65.9 && 66.1 & 66.2 & 67.3 && 68.7 & 68.8 & 71.1 \\
    \midrule
% Knowledge
    \multirow{5}{*}{Knowledge} & $\dagger$AI2D~\cite{kembhavi2016diagram} 
    & 62.0 & 55.5 & 61.1 && 74.8 & 67.0 & 73.8 && 68.9 & 70.1 & 74.3 && 70.8 & 72.0 & 76.3 \\
    & MMMU~\cite{yue2024mmmu} 
    & 32.7 & 35.7 & 33.6 && 38.3 & 41.9 & 39.4 && 34.3 & 39.6 & 39.8 && 35 & 40.4 & 40.5 \\
    & MMStar~\cite{chen2024we}
    & 36.4 & 33.1 & 38.6 && 40.8 & 37.1 & 43.2 && 45.7 & 44.9 & 48.6 && 46.2 & 45.4 & 50.2\\
    & MathVista~\cite{lu2024mathvista} 
    & 30.3 & 25.6 & 30.3 && 42.4 & 35.8 & 42.4 && 46.2 & 44.9 & 44.9 && 52.3 & 50.8 & 50.7 \\
    & $\dagger$ScienceQA~\cite{lu2022learn} 
    & 58.0 & 66.8 & 71.2 && 68.5 & 71.2 & 84.1 && 83.8 & 83.3 & 85.2 && 86.9 & 86.4 & 88.4 \\
    \midrule
    
% Chart and OCR
    \multirow{4}{*}{Chart\&OCR} & $\dagger$ChartQA~\cite{masry2022chartqa} 
    & 39.6 & 31.9 & 40.4 && 46.4 & 37.4 & 47.4 && 49.9 & 51.3 & 51.8 && 51.2 & 52.6 & 53.1 \\
    & $\dagger$TextVQA~\cite{Singh_2019_CVPR} 
    & 44.9 & 45.5 & 44.9 && 51.4 & 52.1 & 51.4 && 53.9 & 54.0 & 53.4 && 54.6 & 54.7 & 54.1 \\
    & OCR-Bench~\cite{liu2023hidden} 
    & 33.6 & 31.8 & 33.9 && 44.8 & 42.5 & 45.2 && 48.7 & 45.5 & 46.7 && 50.2 & 46.9 & 48.1 \\
    & OCRVQA~\cite{mishra2019ocr} 
    & 59.7 & 60.6 & 57.7 && 64.4 & 65.4 & 62.2 && 62.8 & 63.6 & 64.3 && 64.2 & 65.1 & 65.8 \\
    \midrule
% Hallucination
    \multirow{2}{*}{Hallucination}  & POPE~\cite{li2023evaluating} 
    & 87.0 & 86.1 & 84.8 && 87.6 & 87.0 & 85.5 && 87.1 & 86.4 & 86.4 && 87.7 & 87.0 & 87.0 \\
    & HallusionBench~\cite{guan2023hallusionbench} 
    & 30.4 & 27.6 & 32.4 && 30.9 & 28.1 & 33.0 && 34.4 & 32.9 & 35.0 && 33.7 & 35.3 & 37.6 \\
    \midrule
% Vision centric
    Vision Centric & RWQA~\cite{realworldqa2024} 
    & 53.1 & 54.8 & 48.9 && 58.0 & 59.8 & 53.4 && 53.9 & 51.5 & 54.4 && 55.5 & 53.0 & 56.0 \\
    \midrule
    & \multicolumn{1}{r}{\textbf{Average}\mbox{  }} 
    & 47.7 & 48.0 & 48.7 && 54.3 & 53.1 & 55.6 && 55.7 & 55.9 & 57.3 && 56.4 & 56.7 & 58.2 \\
    \bottomrule
    \end{tabular}%
    % }
\caption{\textbf{Evaluation on 15 vision-language benchmarks.} We compare the performance of the regular Panel-of-Peers\pop. We have separated the benchmarks into five categories. Columns show three training iterations for \protect\MIS\ = \pop-Mistral, \protect\VIC\ = \pop-Vicuna, and \protect\LLA\ = \pop-LLaMA3. $\dagger$ indicates that the training set has been observed in our data mixture. For single-try Panel-of-Peers (st-\pop) see the \supp.}
\label{tab:iterations}
\vspace{-0.02in}
\end{table*}

We present three significant results of the \pop methodology, as well as various ablations studying its parameters. The three results comprise (1) a comparison with other state-of-the-art preference optimization methods for LVLMs, (2) the use of \pop as a zero-shot evaluator, and (3) the use of the \pop methodology as a self-improvement algorithm. 

\subsection{Main Results}
Our main result evaluates the efficacy of \pop learning across selected benchmarking metrics, showcasing its superior performance against a competitive set of preference alignment models. For fairness, we keep the original \texttt{LLaVA-1.5} configuration using the \mbox{Vicuna-7B} language model and CLIP/L-14 vision model, as well as, comparing only to \pop performed for only one iteration. That way, we ensure that the data used stays in the same order of magnitude as previous methods.
As depicted in Table~\ref{tab:comparission}, our method consistently outperforms other state-of-the-art preference alignment methods on these benchmarks. Notably, \pop-iter1 achieves a score of 68.7\% on the MMbench, 67.9\% on the SEED-Bench, and 35.6\% on the MM-Vet, demonstrating its robust capabilities in complex multi-modal scenarios. Furthermore, its score of 71.2\% on ScienceQA highlights the ability of our method to improve performance on scientific question-answering tasks. These results show the effectiveness of our strategy as a simple yet powerful way to create data for self-improvement.

\subsection{Panel-of-Peers as a Zero-Shot Evaluator}
\label{subsec: zero-shot-pop}
In this experiment, we evaluate the \pop algorithm’s capacity as a zero-shot evaluator by creating panels at different model scales and by sampling 15 candidate responses from each panel member and harnessing the ensemble’s collective judgment to select the best answer.
We constructed panels at four distinct scales by selecting different sets of models based on their parameter count: $>3$B, $>7$B, $>10$B, and $>30$B. At each scale, the selected models not only generate answers but also evaluate their own responses and those from the rest of the panel. As shown in Figure~\ref{fig:vlmreward}, the st-\pop and \pop frameworks consistently outperform the average single-model approach (Avg-Single). This baseline represents the average score across all individual models within a panel, serving as a point of comparison. Our results demonstrate that the ensemble scores for \pop and st-\pop outperform the baseline at every model scale. This suggests that a peer-assessment process enhances answer quality by aggregating judgments across diverse models, effectively creating an ensemble that is more accurate than any individual contributor. Details of the specific models used for this experiment can be found in the \supp.

\subsection{Self-Improvement from a Panel-of-Peers}
\begin{figure*}[t!]
    \centering
    \begin{minipage}{0.31\textwidth}
        \centering
        \includegraphics[width=\linewidth]{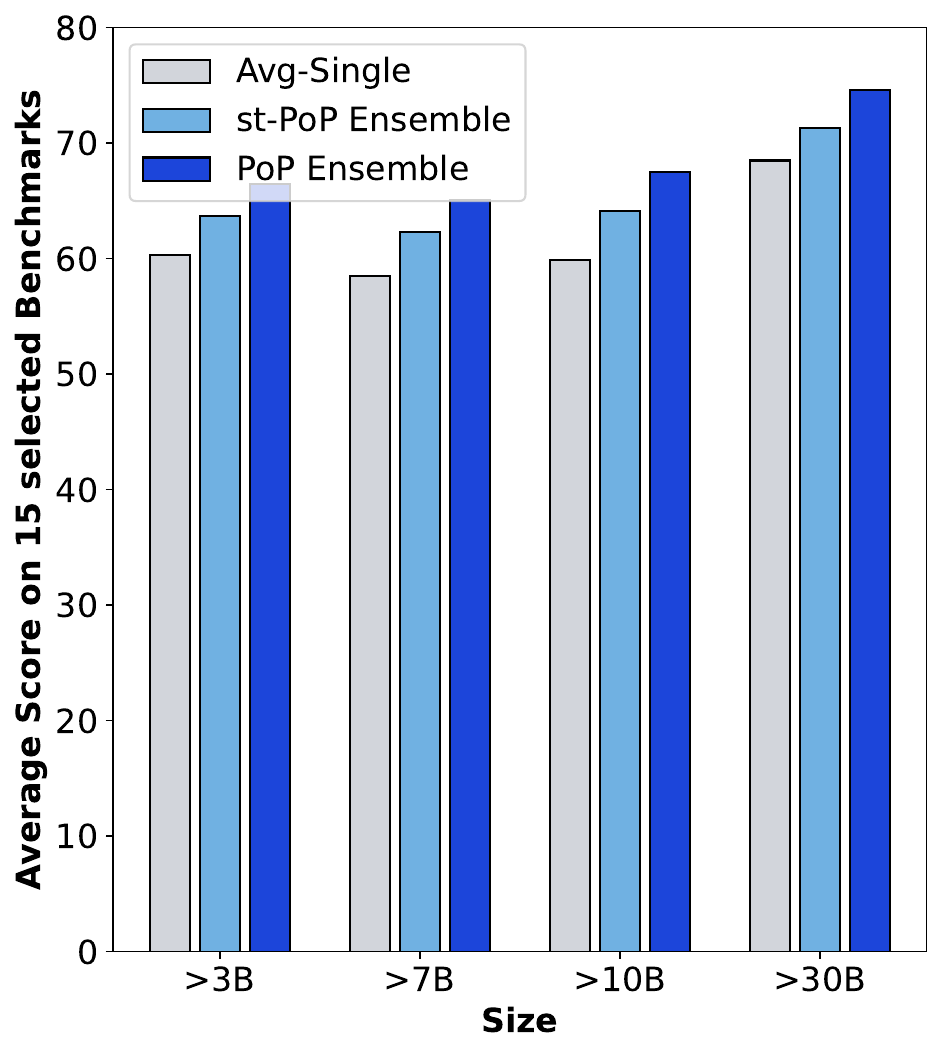}
        \caption{\textbf{Panel-of-Peers as a reward model.} Average scores of the Panel-of-Peers used as a reward model on 15 selected benchmarks}
        \label{fig:vlmreward}
        \vspace{5mm} % Adjust this value to align the top
    \end{minipage}
    \hfill
    \begin{minipage}{0.31\textwidth}
        \centering
        \includegraphics[width=\linewidth]{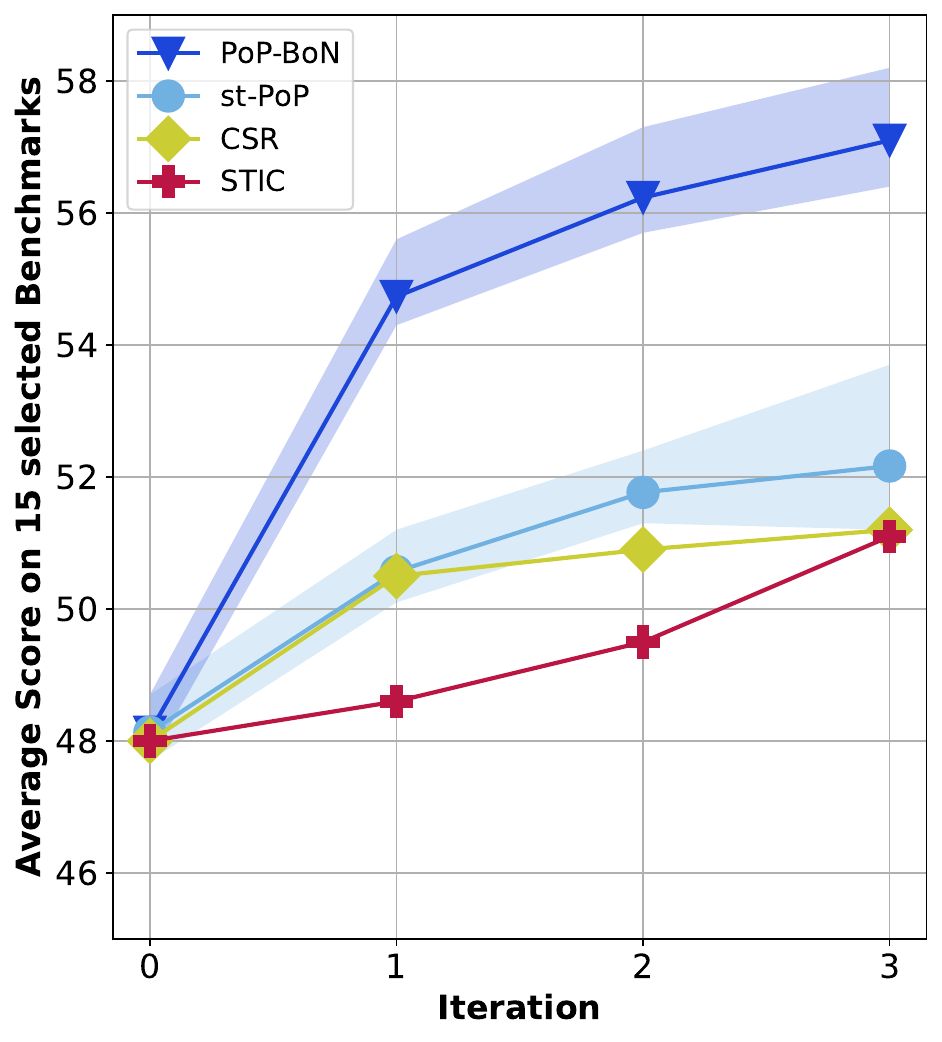}
        \caption{\textbf{Self-improvement iterations.} Average scores of PoP learning at different iterations of self-improvement over 15 selected benchmarks}
        \label{fig:iterations}
        \vspace{5mm} % Adjust this value to align the top
    \end{minipage}
    \hfill
    \begin{minipage}{0.32\textwidth}
        \centering
        \includegraphics[width=0.95\linewidth]{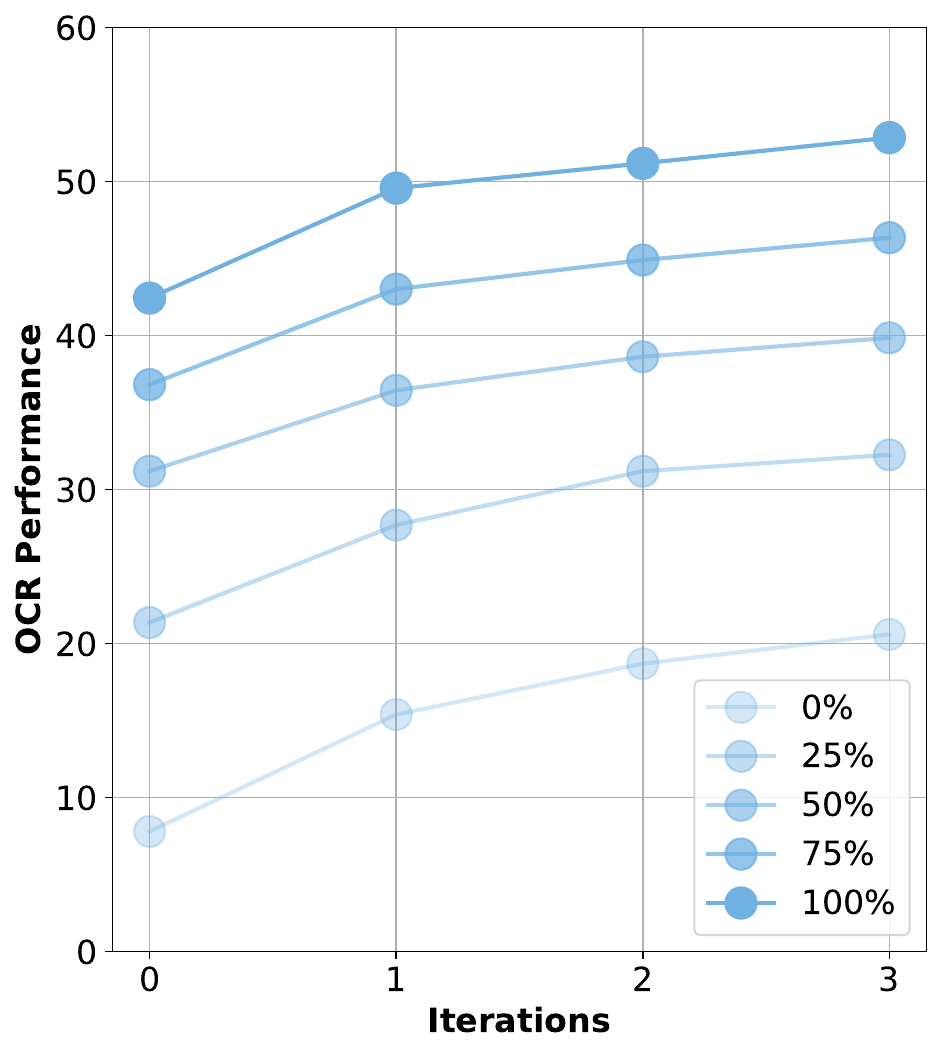}
        \caption{\textbf{Learning a new skill from peers} We start with a model with a limited knowledge of OCR ($\approx$0\% - 100\%) and use \pop to teach the model OCR knowledge.}
        \label{fig:learn_ocr}
        \vspace{5mm} % Adjust this value to align the top
    \end{minipage}
\vspace{-0.25in}
\end{figure*}

As illustrated in Figure~\ref{fig:iterations}, our method, \pop, consistently improves its performance over multiple iterations of self-improvement, achieving higher average scores on the 15 selected benchmarks compared to CSR~\cite{zhou2024calibrated} and STIC~\cite{deng2024enhancing}. For fairness, we keep the original \texttt{LLaVA-1.5} configuration using the \mbox{Vicuna-7B} language model and CLIP/L-14 vision model. \pop exhibits a steady upward trajectory, outperforming the other methods as the number of iterations increases, but seems to plateau after the third iteration. 

\noindent We report the per-iteration performance of each student in the \pop, on 15 vision-language benchmarks grouped into five categories (General VQA, Knowledge, Chart \& OCR, Hallucination, and Vision Centric). See Table~\ref{tab:iterations}. Each member in the panel is initialized with the model at iteration 0. The table illustrates how iterative self-improvement consistently boosts performance across all benchmarks, highlighting the benefits of our peer-feedback approach.

\subsection{Learning an Ability from Peers.}
In this experiment, we test whether peer-to-peer learning can unlock a new ability, not just improve an existing ability. To do this, our panel includes an additional model, which we designed to have minimal OCR capabilities. We refer to this model as \emph{OCR-Dumb}. With this new four-member panel, we implemented a self-improvement loop to assess whether the ``student'' models could elevate the OCR-Dumb model performance to match its peers. Furthermore, we investigated how this improvement impacted other tasks such as general knowledge, math, hallucination detection, and vision-centric.

We initiate the OCR-Dumb model with progressively greater amounts of OCR knowledge, simulated by using 0\%, 25\%, 50\%, 75\%, and finally 100\% of available OCR data in training. Each level of OCR knowledge was evaluated across five general categories: Chart \& OCR, General VQA, Knowledge, Hallucination, and Vision-Centric tasks, as shown in Figure~\ref{fig:learn_ocr}. As the OCR-Dumb model improved its reading capabilities, we observed a steady increase in performance across all categories. This indicates that OCR knowledge is a crucial component for reading-specific tasks and for tasks requiring an understanding of structured visual data or answering knowledge-based questions involving text. Interestingly, the model's improvement in hallucination detection and vision-centric tasks suggests that enhanced OCR capabilities contribute to better general alignment in multimodal understanding. This experiment shows how Peer-to-Peer Learning, with varying levels of specialized knowledge, can iteratively improve a model’s core abilities, even in areas where it initially struggles.

\subsection{Ablations}
\label{sec:ablations}

\begin{figure*}[t!]
  \centering
\begin{subfigure}{0.24\textwidth}
    \centering
    \includegraphics[width=\linewidth]{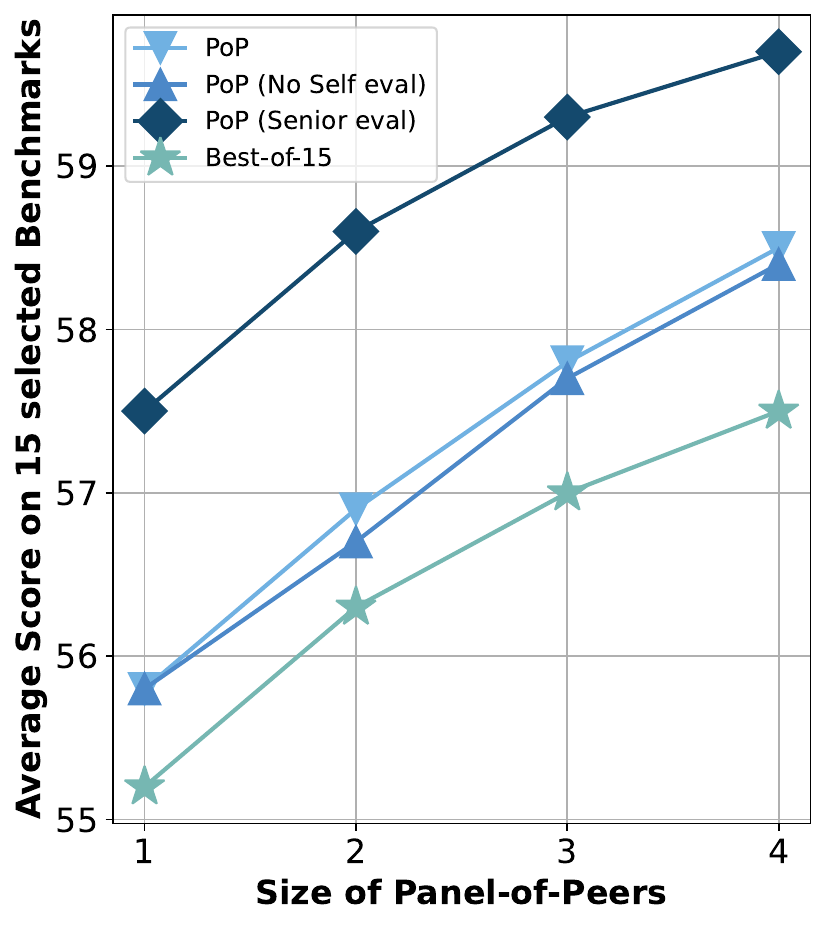}
    \caption{\footnotesize{\textbf{Types of Panel-of-Peers.}}}
    \label{fig:best_panel}
  \end{subfigure}
\begin{subfigure}{0.24\textwidth}
    \centering
    \includegraphics[width=\linewidth]{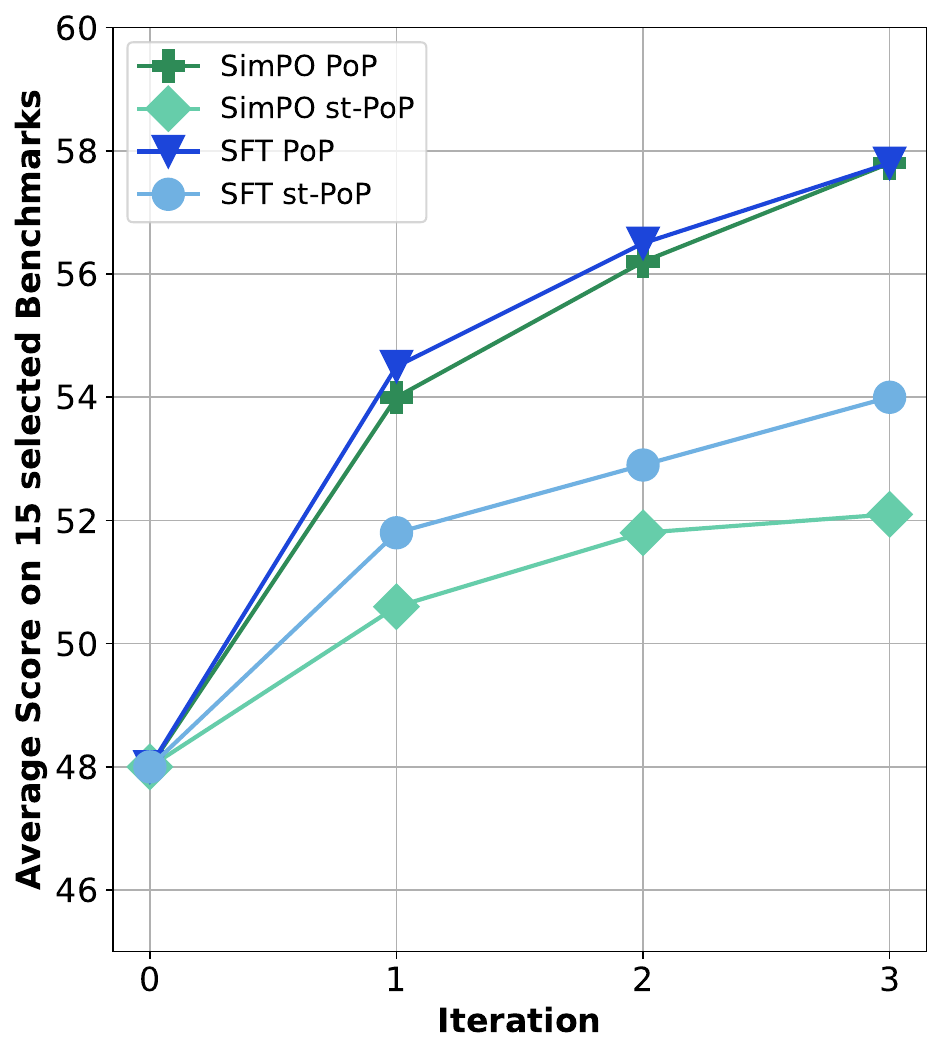}
    \caption{\footnotesize{\textbf{Alignment objective.}}}
    \label{fig:dpo_sft}
  \end{subfigure}
  \begin{subfigure}{0.24\textwidth}
    \centering
    \includegraphics[width=\linewidth]{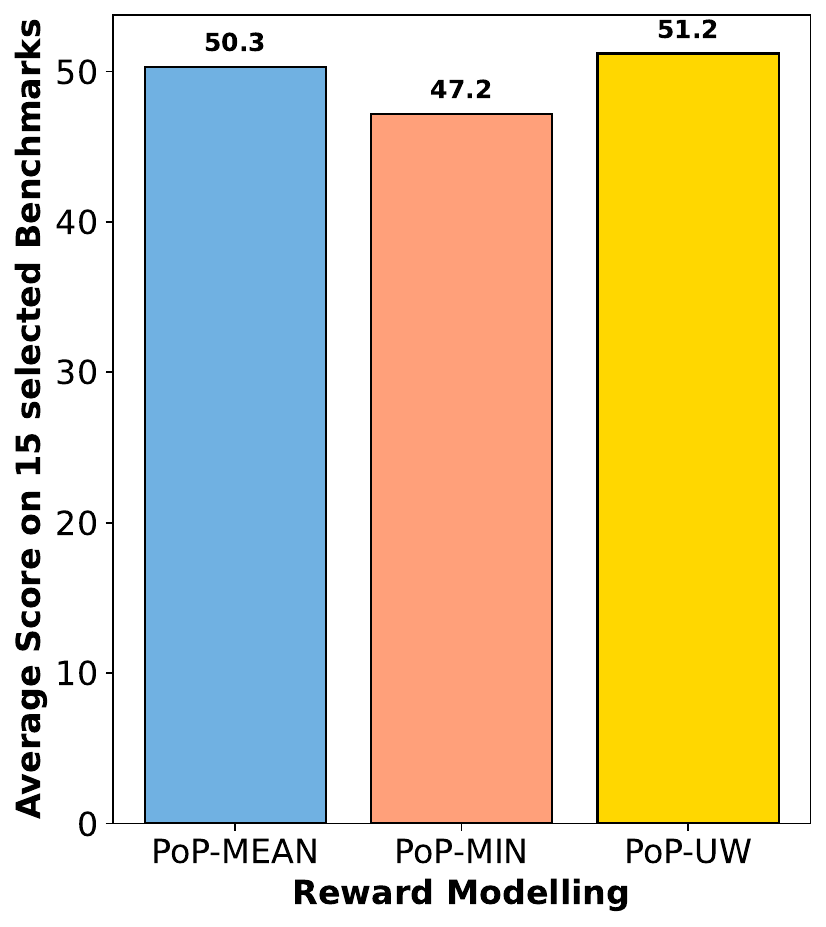}
    \caption{\footnotesize{\textbf{Reward aggregation methods.}}}
    \label{fig:reward}
  \end{subfigure}
  \begin{subfigure}{0.24\textwidth}
    \centering
    \includegraphics[width=\linewidth]{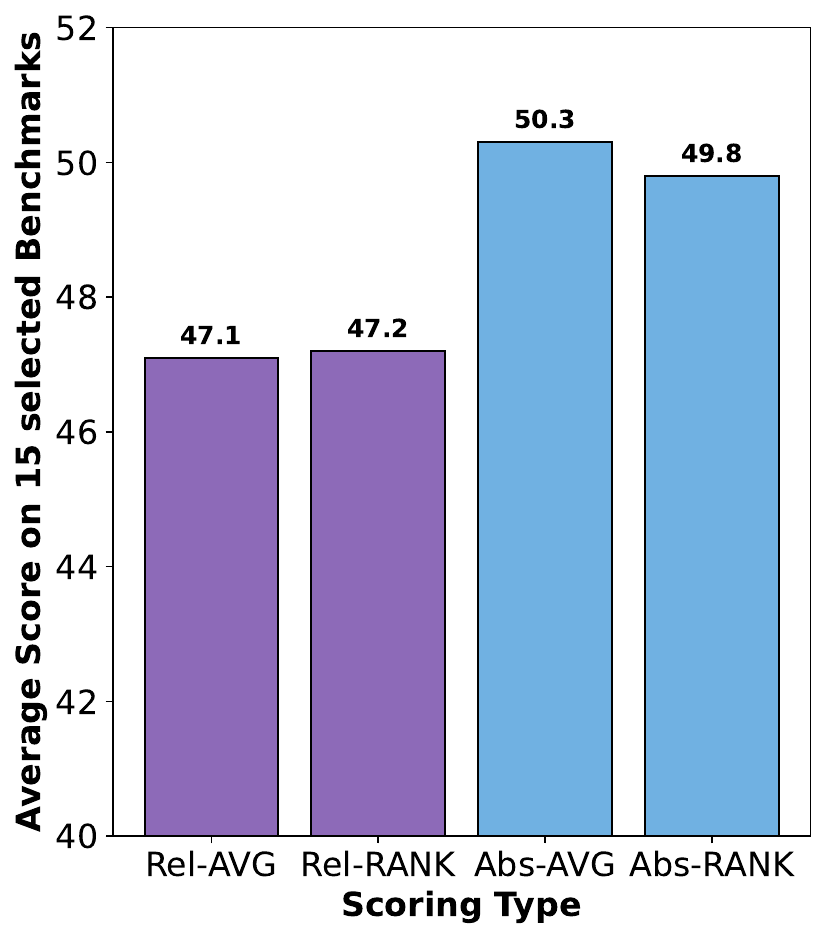}
    \caption{\footnotesize{\textbf{Prompt scoring methods}}}
    \label{fig:prompt_score}
  \end{subfigure}

  \caption{\textbf{Peer-to-Peer Learning ablation experiments} We use the \texttt{LLaVA-1.5}, configuration using the Vicuna-7B language model and CLIP/L-14 vision model except for the first ablation which is evaluation only. We evaluate on 15 selected benchmarks.}
  \label{fig:ablations}
\vspace{-0.1in}
\end{figure*}

\noindent{\bf What makes a good panel?}
A good panel in PoP Learning combines diversity, unbiased evaluation, and iterative feedback. In the standard PoP setup, each model in the panel assesses the responses of others, promoting a collaborative, classroom-like environment where models learn from their peers. Excluding self-evaluation (PoP - No Self Eval) has minimal impact, likely because the evaluations are blind and do not introduce self-bias. In the "Senior Eval" setup, however, smaller models (e.g., \pop-Vicuna, \pop-Mistral, \pop-LLaMA3) generate answers while larger models (LLaVA-NeXT-34B~\cite{liu2024llavanext}, InternVL2-40B~\cite{chen2024far}) grade them. This setup boosts overall scores by leveraging ``senior'' models as reviewers but lacks the iterative learning benefits of having all models serve as both generators and evaluators, which enables continuous refinement across iterations. Finally, in the Best-of-15 configuration, a single model (\pop-Vicuna) generates multiple responses, and the panel (\pop-Vicuna, \pop-Mistral, \pop-LLaMA3) evaluates and selects the best answer. This setup underperforms compared to diverse panels, likely because it reduces exposure to varied perspectives, focusing on optimizing a single model's responses rather than leveraging the strengths of multiple models. These results can be seen in detail in Figure~\ref{fig:best_panel}.

\noindent{\bf Aligment objective.} 
We examine the effect of the alignment objective on performance by comparing two strategies: Supervised Fine-Tuning (SFT) and Simple Preference Optimization (SimPO)~\cite{meng2024simpo}. For this experiment, we construct an SFT dataset by selecting the best from the panel. This allows us to create data $\mathcal{D} = \{(\mathcal{I}, \mathbf{x}, \mathbf{y}^{(m)})\}_{i=1}^{|\mathcal{D}|} $ with only positive feedback from the panel. As shown in Figure~\ref{fig:dpo_sft}, our results indicate that using SFT with the Panel-of-Peers framework is match with SimPO. When we use a single try per model, the reward margin in this setup is relatively limited, providing less separation for preference-based optimization to exploit. SFT’s structured approach to fine-tuning appears more effective in this scenario, likely due to training directly on answers. Interestingly, when we allow multiple attempts per model, the performance gap between SFT and SimPO diminishes. With \pop, models are given up to 15 attempts to respond, increasing the likelihood of a larger margin between the best and worst responses. This bigger margin allows SimPO to benefit more from the distinct variations in quality, thereby improving its ability to optimize based on preference data. Finally, SFT, while on average matched with SimPO, essentially reduces to knowledge distillation with rejection sampling~\cite{li2024multi}. SFT does not lead to reductions in hallucinations or improvements in vision-centric tasks; we believe that the rejected answers help steer the model away from hallucinations—a benefit lost when training solely on the preferred answer.

\noindent{\bf Relative vs absolute scoring.}
 The Relative Scoring approach allows each model to evaluate all responses simultaneously, assigning either a relative score or ranking to each, which can then be aggregated by averaging or by majority vote. This method is computationally efficient as it enables faster evaluations. However, the Absolute Scoring approach, where each model evaluates responses individually without comparing them to others, demonstrated better performance. Absolute scoring can be aggregated in a similar way. As seen in the figure, Absolute scoring with averaging produced the highest average score, followed closely by Absolute scoring with ranking, indicating that, despite being slower, absolute scoring yields more reliable evaluations than relative scoring due to its focused assessment of each response. Though costlier, it can be parallelized and improved with advances in model inference. In \pop (three models generating 45 answers), relative scoring prompts become excessively long, causing {\em loss of context}.

\noindent{\bf Reward modeling.}
We evaluated three strategies for aggregating scores from the panel members. \pop-MEAN, involves a straightforward averaging of scores across panel members, offering simplicity and ease of implementation. \pop-MIN adopts a conservative approach by selecting the minimum score, ensuring that at least one model perceives the response as sufficiently accurate to pass. Finally, \pop-UW (Uncertainty-Weighted), incorporates an uncertainty-aware mechanism that weights scores based on inter-model variance, adjusting for potential inconsistencies in panel evaluations. As shown in the figure, the uncertainty-weighted approach achieves the highest average score, followed closely by averaging, while the conservative approach lags slightly behind. However, due to its practical simplicity, we opted for averaging as the primary method for reward aggregation.

\noindent{\bf Does \pop use extra data?}
The 900K images and queries from Cambrian-7M used to generate preference data were not part of the initial training data. However, it is important to emphasize that we only used the queries and images, excluding the ground truth answers, during the self-improvement process.
\noindent
We hypothesized that a panel of peers can iteratively improve their collective understanding through feedback, similar to students completing end-of-chapter exercises. This hypothesis was validated by our results. For completeness,  we performed supervised fine-tuning on the full 900K ground truth samples (corresponding to three iterations of self-improvement). The LLaVA-Vicuna model fine-tuned on this ground truth data achieved an average score of 54.0, 3\% below the 57.0 achieved by \pop-Vicuna. These findings reinforce the effectiveness of \pop, demonstrating that using peer-generated feedback can produce high-quality synthetic data, which in turn facilitates learning that surpasses what is achieved with static, short, and noisy annotated datasets.

\section{Discussion and Conclusion}
\label{sec:conclusion}
We introduced Panel-of-Peers (PoP) learning, a new self-improvement approach for enhancing LVLMs, where we showed that a panel of models with similar initial capabilities can be used as both candidate response generator and evaluators to synthesize high-quality and diverse data, which 
enables us to iteratively and independently enhance the performance of individual panel members over multiple rounds.

Our results from an extensive set of benchmarks and ablations clearly show that an iterative improvement can be achieved by simply working on a dataset of queries (\eg PoP-LLaMA3 model surpasses their LLaVA counterparts by 9 absolute points), which can reduce or eliminate the need to collect expensive human annotations.

The PoP approach  
can also address specific weaknesses by leveraging complementary strengths in the panel. For example, a model with limited OCR abilities can benefit from peer models with stronger OCR capabilities, fostering cross-model knowledge transfer. The PoP framework imposes no limitations on model size or capacity as long as they are similar performing models, allowing PoP to scale seamlessly with future frontier models and making it a versatile tool for the rapid evolution of LVLMs.   
Due to its design, PoP is also easily extensible. Better contrastive alignment methods, alternate sampling methods such as beam search variants, and other advancements in LVLMs can be independently useful alongside our PoP method. 

\vspace{0.05in}
\noindent {\bf Acknowledgments. } V. Ordonez is funded by an NSF CAREER Award \#2201710 and the Ken Kennedy Institute. 

{\small
\bibliographystyle{ieeenat_fullname}
\bibliography{sections/11_references}
}

\ifarxiv \clearpage \appendix \appendix
\counterwithin{figure}{section}
\counterwithin{table}{section}
\renewcommand\thefigure{\thesection.\arabic{figure}}
\renewcommand\thetable{\thesection.\arabic{table}} 

\section{Implementation Detatils}

\begin{figure*}[t!]
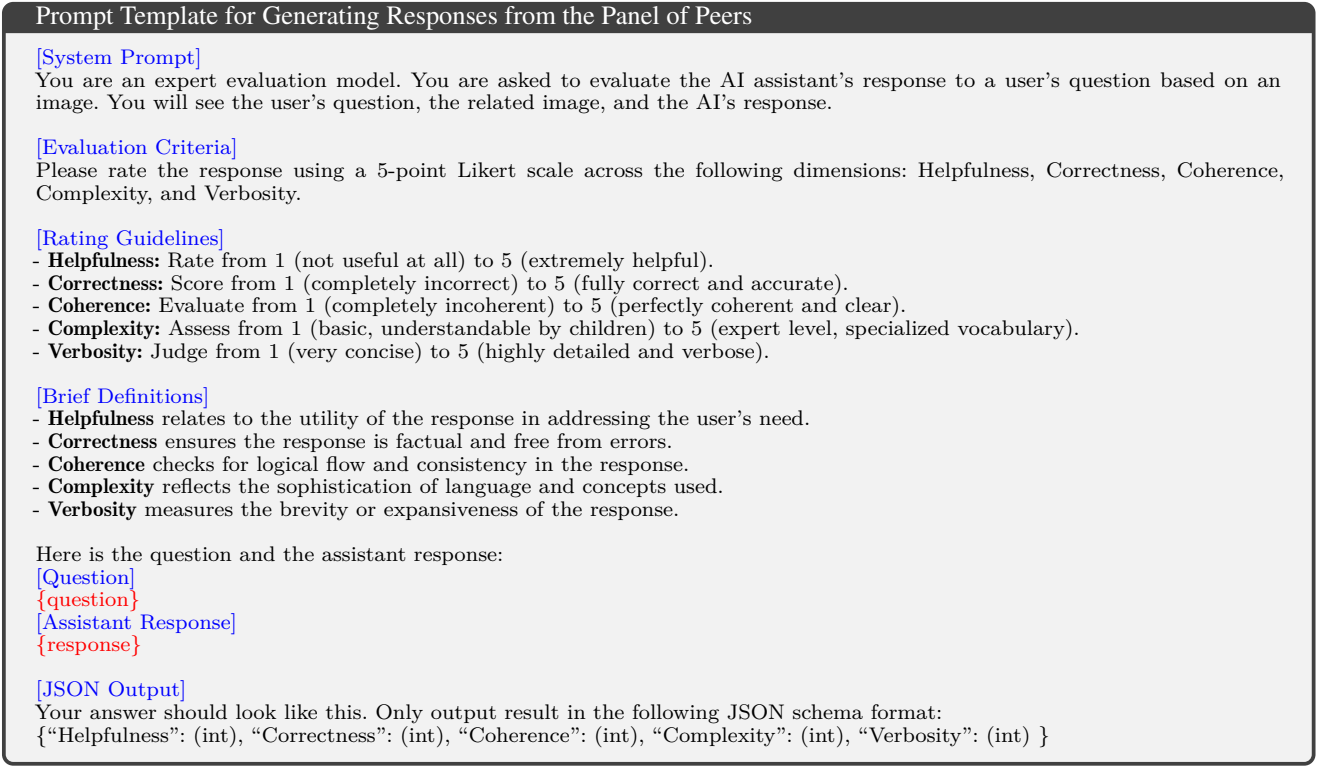

    \centering
    \begin{prompt}{Prompt Template for Generating Responses from the Panel of Peers}
{\color{blue}[System Prompt]}
\\
You are an expert evaluation model. You are asked to evaluate the AI assistant's response to a user's question based on an image. You will see the user's question, the related image, and the AI's response.
\\
\\\relax
{\color{blue}[Evaluation Criteria]}
\\
Please rate the response using a 5-point Likert scale across the following dimensions: Helpfulness, Correctness, Coherence, Complexity, and Verbosity.
\\
\\\relax
{\color{blue}[Rating Guidelines]}
\\
- \textbf{Helpfulness:} Rate from 1 (not useful at all) to 5 (extremely helpful). \\
- \textbf{Correctness:} Score from 1 (completely incorrect) to 5 (fully correct and accurate). \\
- \textbf{Coherence:} Evaluate from 1 (completely incoherent) to 5 (perfectly coherent and clear). \\
- \textbf{Complexity:} Assess from 1 (basic, understandable by children) to 5 (expert level, specialized vocabulary). \\
- \textbf{Verbosity:} Judge from 1 (very concise) to 5 (highly detailed and verbose).
\\
\\\relax
{\color{blue}[Brief Definitions]}
\\
- \textbf{Helpfulness} relates to the utility of the response in addressing the user's need. \\
- \textbf{Correctness} ensures the response is factual and free from errors. \\
- \textbf{Coherence}  checks for logical flow and consistency in the response. \\
- \textbf{Complexity} reflects the sophistication of language and concepts used. \\
- \textbf{Verbosity} measures the brevity or expansiveness of the response.
\\
\\
Here is the question and the assistant response:
\\\relax
{\color{blue}[Question]} \\
{\color{red}\{question\}} \\\relax
{\color{blue}[Assistant Response]} \\
{\color{red}\{response\}}
\\
\\\relax
{\color{blue}[JSON Output]}
\\
Your answer should look like this. Only output result in the following JSON schema format: \\
\{``Helpfulness'': (int), ``Correctness'': (int), ``Coherence'': (int), ``Complexity'': (int), ``Verbosity'': (int) \}
\end{prompt}
    \caption{{\bf Evaluating Synthetic Responses.} We use the following prompt template, which is used to evaluate responses from the Panel-of-Peers.}
    \label{fig:prompt}
\end{figure*}

\subsection{Training Hyperparameters}
In Table~\ref{tab:hyperparameters}, we list the detailed training dataset usage and hyperparameters. The training data are constructed based on the following datasets: \texttt{BLIP-LAION-CC-SBU}~\cite{liu2024visual}, which contains 558K image-text pairs from BLIP-captioned CC3M~\cite{li2022blip}, SBU~\cite{ordonez2011im2text}, and LAION400M~\cite{schuhmann2021laion} filtered by LLaVA; \texttt{LLaVA-Instruct-mix665k}~\cite{liu2023improvedllava}, which contains 665k visual instruction-following data constructed to train the LLaVA family of models; and synthetic data created using images and questions from the \texttt{Cambrian-7M} dataset~\cite{tong2024cambrian}. Unless otherwise specified, we randomly sample the indicated number of instances from each dataset during the training process. During training, we use Flash Attention~\cite{dao2022flashattention}, bfloat16, and PyTorch FSDP~\cite{zhao2023pytorch} to accelerate training efficiency.

\subsection{Panel-of-Peers Models}
\noindent{\bf Image Processing and Visual Representations} 
We implement all image processing logic using the default image transforms provided by \texttt{torchvision} and the TIMM library~\cite{rw2019timm}. We normalize pixel values using the default ImageNet normalization values. The default backbone employed by all visual representations that we evaluate in this work is a Vision Transformer~\cite{dosovitskiy2020image} trained with the CLIP objective~\cite{radford2021learning}; we extract patch features from the \textit{penultimate} layer, following LLaVA~\cite{liu2024visual}.

\noindent{\bf Vision-Language Projector} 
We use a simple 2-layer GELU MLP as the projector, which projects each patch independently into the embedding space of the language model.

\noindent{\bf Language Model} 
We choose three models to create the Panel-of-Peers: Vicuna-7B~\cite{vicuna2023}, Mistral-7B~\cite{jiang2023mistral}, and -8B~\cite{llama3model}. In order to combine the projected visual patch embeddings, we perform simple sequence-wise concatenation by placing the patch embeddings before the text embeddings.

\subsection{Evaluation benchmarks}
Systemic evaluations of the Panel-of-Peers regarding General VQA, knowledge, Chart\&OCR, Hallucination, and Vision-Centric capabilities have been conducted. The benchmarks and datasets used are listed in Table~\ref{tab:eval_benchmarks_summary}. During the evaluation, we use \texttt{VLMEvaKit}~\cite{duan2024vlmevalkit} as our primary evaluation toolkit.

\begin{table*}[t!]
\label{tab:details}
\footnotesize
\centering
% \resizebox{\linewidth}{!}{
\begin{tabular}{lccc}
 & \texttt{Stage I} & \texttt{Stage II} & \texttt{Stage III}\\
 \toprule[0.95pt]
 Config & Alignment & SFT & PoP\\
 \midrule[0.6pt]
 \multicolumn{4}{c}{\textit{Training Hyper-Parameters}} \\
 \midrule[0.6pt]
 Optimizer & AdamW & AdamW & AdamW\\
 Learning Rate & 2e-3 & 2e-5 & 6e-5\\
 Weight Decay & 0.0 & 0.0 & 0.0\\
 Training Epochs & 1 & 1 & 2\\
 Warmup Ratio & 0.003 & 0.003 & 0.003 \\
 Learning Rate Scheduler & Cosine & Cosine & Cosine\\
 Batch Size Per GPU & 16 & 8 & 8\\
 Maximum Token Length & 2048 & 2048 & 2048\\
 Unfreeze LLM & \xmark & \cmark & \cmark \\
 \midrule[0.6pt]
 \multicolumn{4}{c}{\textit{Training Data}} \\
 \midrule[0.6pt]
Dataset & \texttt{BLIP-LAION-CC-SBU} &  \texttt{LLaVA-Instruct-mix665k} & Sampled from \texttt{Cambrian-7M} \\
 \midrule[0.3pt]
Data Size & 558K & 665K & 3 $\times$300K \\
 Data Type & Pair & Instruction & Synthetic\\
  \midrule[0.6pt]
 \multicolumn{4}{c}{\textit{Training Cost}} \\
 \midrule[0.6pt]
 GPU Device & 8$\times$NVIDIA A100-80GB & 8$\times$NVIDIA A100-80GB & 8$\times$NVIDIA A100-80GB\\ 
 Training Time & $\sim$6h & $\sim$10h & $\sim$90h\\
\bottomrule[0.95pt]
\end{tabular}
% }
\caption{\textbf{Training recipes} for \pop. The three training stages are introduced in Section 3. \texttt{Stage I}: Alignment training, \texttt{Stage II}: Instruction Tuning, \texttt{Stage III}: Panel-of-Peers Learning.}
\label{tab:hyperparameters}
\end{table*}

\begin{table*}[t]
\centering
% \tiny
\setlength{\tabcolsep}{4pt}
\renewcommand{\arraystretch}{1.2}
% \resizebox{\textwidth}{!}{%
\begin{tabular}{@{}lllll@{}}
\toprule[0.95pt]
Capability & Dataset & Task description & Eval Split & Metric \\
\midrule
\multirow{3}{*}{General VQA} 
& MM-Vet~\cite{yu2024mmvet}  & Multi-disciplinary QA   & - &  GPT-4 Eval~\cite{yu2024mmvet} \\
& MMBench~\cite{liu2023mmbench}& Multi-disciplinary QA   & \texttt{dev} &  GPT-3.5 Eval~\cite{liu2023mmbench} \\
& SEED-Bench~\cite{li2024seed}& Multi-disciplinary QA & -  &   Multi-choice Acc     \\
\midrule
 \multirow{5}{*}{Knowledge}
& AI2D~\cite{kembhavi2016diagram} & Science Diagrams  &  \texttt{test}   & Multi-choice Acc\\
& MMMU~\cite{yue2024mmmu} & College-level Multi-disciplinary  & \texttt{val} & Multi-choice Acc\\
& MMStar~\cite{chen2024we} & Misc Multi-disciplinary  & - & Multi-choice Acc\\
& MathVista~\cite{lu2024mathvista} & General Math Understanding  & \texttt{min} & GPT-4 Eval\\
& ScienceQA~\cite{lu2022learn} & High-school Science  & \texttt{val} & Multi-choice Acc\\
\midrule
\multirow{4}{*}{Chart\&OCR}
& ChartQA~\cite{masry2022chartqa} & Chart Understanding & \texttt{test} & Relaxed Accuracy\\
& TextVQA~\cite{Singh_2019_CVPR}  &  OCR; Reasoning &  \texttt{val}   &  VQAScore\\
& OCR-Bench~\cite{liu2023hidden} & OCR; Multi-disciplinary & -& Acc\\
& OCRVQA~\cite{mishra2019ocr}  & Document OCR & \texttt{TESTCORE} & Acc\\
\midrule
\multirow{2}{*}{Hallucination}
& POPE~\cite{li2023evaluating}  & Yes/No Hallucinations & - & Acc, F1-score\\
& HallusionBench~\cite{guan2023hallusionbench} & Visual Hallucination & - & Acc, F1-score\\
\midrule
Vision Centric
& RWQA~\cite{realworldqa2024} & Real-world QA & \texttt{dev} & Multi-choice Acc\\
\bottomrule
\end{tabular}
% }
\caption{\textbf{Overall descriptions of the evaluation benchmarks} for evaluating capabilities, including GeneralVQA, Knowledge, Chart\&OCR, Hallucination and Vision Centric Benchmarks.}
\label{tab:eval_benchmarks_summary}
\end{table*}

\subsection{Prompt Template}
To evaluate model-generated responses within our Panel-of-Peers (PoP) framework, we designed a detailed prompt template to guide models in rating responses. This prompt was central to generating pseudo-rewards, which serve as feedback signals to enable self-improvement iterations. Each model evaluated the outputs of its peers based on a set of predefined criteria and aggregated their results using an ensemble strategy to achieve consensus. The prompt comprises three main components: \textit{System Prompt}, \textit{Evaluation Criteria}, and \textit{Rating Guidelines}. It is structured as follows:
\begin{itemize}
    \item \textbf{System Prompt:} The model is instructed to act as an expert evaluator tasked with assessing the quality of a response provided to a user's question. Both the question and its related image are provided for context.
    \item \textbf{Evaluation Criteria:} Responses are evaluated across five dimensions on an ordinal Likert scale:
    \begin{enumerate}
        \item \textbf{Helpfulness:} Utility of the response in addressing the user's query (1 to 5 scale).
        \item \textbf{Correctness:} Accuracy and factuality of the response (1 to 5 scale).
        \item \textbf{Coherence:} Logical consistency and clarity of the response (1 to 5 scale).
        \item \textbf{Complexity:} Level of language sophistication, ranging from simple to expert-level (1 to 5 scale).
        \item \textbf{Verbosity:} Appropriateness of detail and conciseness (1 to 5 scale).
    \end{enumerate}
    \item \textbf{Rating Guidelines:} Models receive detailed explanations for scoring each dimension. For instance, a rating of 5 in Helpfulness indicates complete alignment with the user's intent, while a 1 represents a failure to address the query effectively. Similarly, Coherence is rated based on logical flow, with a 1 indicating substantial contradictions or redundancy.
    \item \textbf{Output Format:} To standardize results, models are instructed to provide evaluations in a strict JSON schema format, including scores for each criterion.
\end{itemize}
This prompt enabled consistent and systematic evaluation of the model-generated responses, ensuring that pseudo-rewards were aligned with the evaluation objectives outlined in our PoP framework.

\begin{figure*}[t!]
    \centering
    \includegraphics[width=\textwidth]{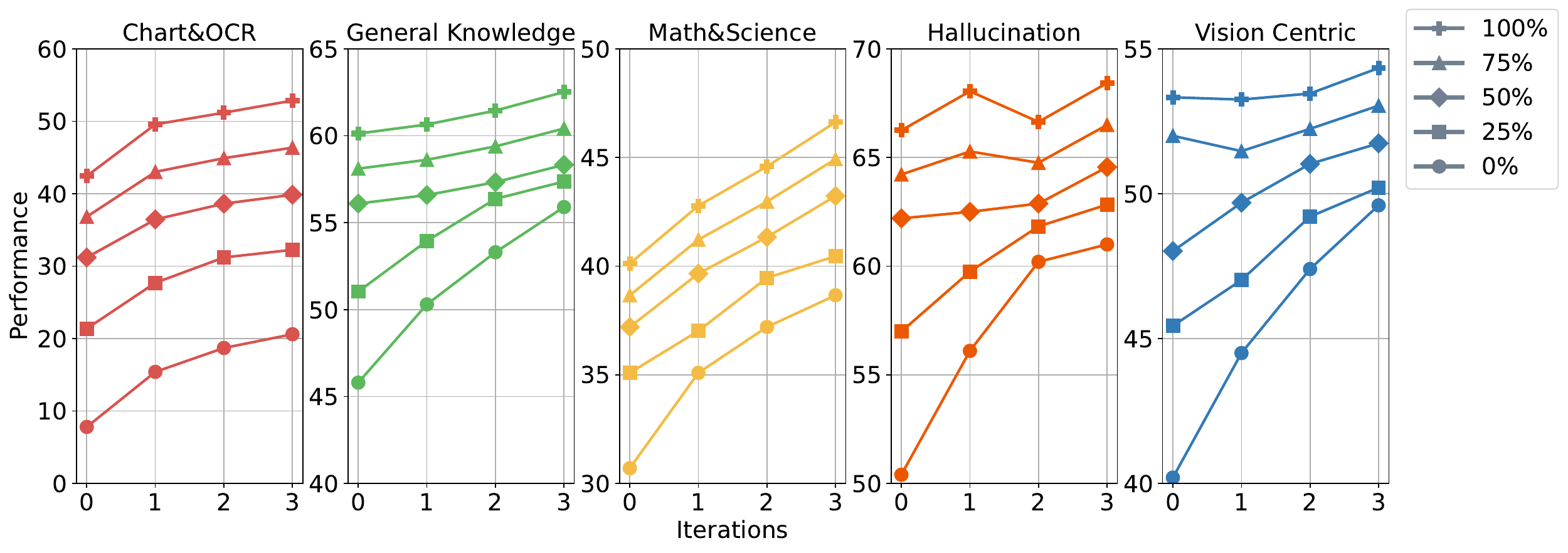}
    \caption{\textbf{Learning a New Skill from Peers (OCR).} We start with a model with very limited OCR knowledge ($\approx 0\%$) and use PoP to iteratively teach OCR skills. The performance is evaluated across multiple categories, including Chart \& OCR, General Knowledge, Math \& Science, Hallucination, and Vision-Centric tasks.}
    \label{fig:learn_ocr_full}
    \vspace{-0.2in}
\end{figure*}

\section{Additional Experiments}
\subsection{Comparison with State of the Art}
We compare against the top 49 models on the OpenVLM leaderboard, highlighting the performance of our models using \pop. Our models include \pop-Vicuna, \pop-Mistral, \pop-LLaMA3, and their single-try counterparts, which are evaluated in 15 benchmarks against a broad spectrum of state-of-the-art methods.

\noindent
Our best-performing models, \pop-LLaMA3 and mt-\pop-LLaMA3, achieve an average score of 56.3\% and 59.7\%, starting from a score of 48.9\%. Compared to proprietary models like GPT4-o~\cite{openai_gpt4o_system_card_2024} and Gemini-1.5~\cite{team2023gemini}, our models lag behind approximately 20 percentage points in performance. A similar gap is observed when compared with open-source state-of-the-art models, such as Qwen2-VL-72B~\cite{Qwen2VL}, InternVL2-Llama3-76B~\cite{chen2024far}, and NVLM-D-72B~\cite{nvlm2024}. Compared to models of the same size category but trained on significantly more data and higher-resolution inputs, our best-performing models lag behind the recently released Qwen2-VL-7B~\cite{Qwen2VL}, the LLaVA-OneVision family~\cite{li2024llava}, and the Molmo family~\cite{deitke2024molmo} by approximately 10 percentage points. Compared to models of the same size category trained on similar budgets, our best-performing model surpasses all the LLaVA-NeXT family~\cite{liu2024llavanext} except for models larger than 30B by approximately 5 percentage points. We remark that our models use 224x224 pixels as the input resolution compared to 768x768 pixels of the NeXT family.

These results demonstrate the efficacy of our approach in using peer evaluations to improve model performance, effectively increasing the average score by approximately 12\% compared to the original \texttt{LLaVA-1.5-7b} model. Figure~\ref{fig:comparisions} illustrates a comparative analysis of the top 49 models on the OpenVLM leaderboard~\cite{duan2024vlmevalkit}, highlighting the performance of our models using peer-to-peer learning.

\begin{figure*}[t!]
    \centering
    \includegraphics[width=\textwidth]{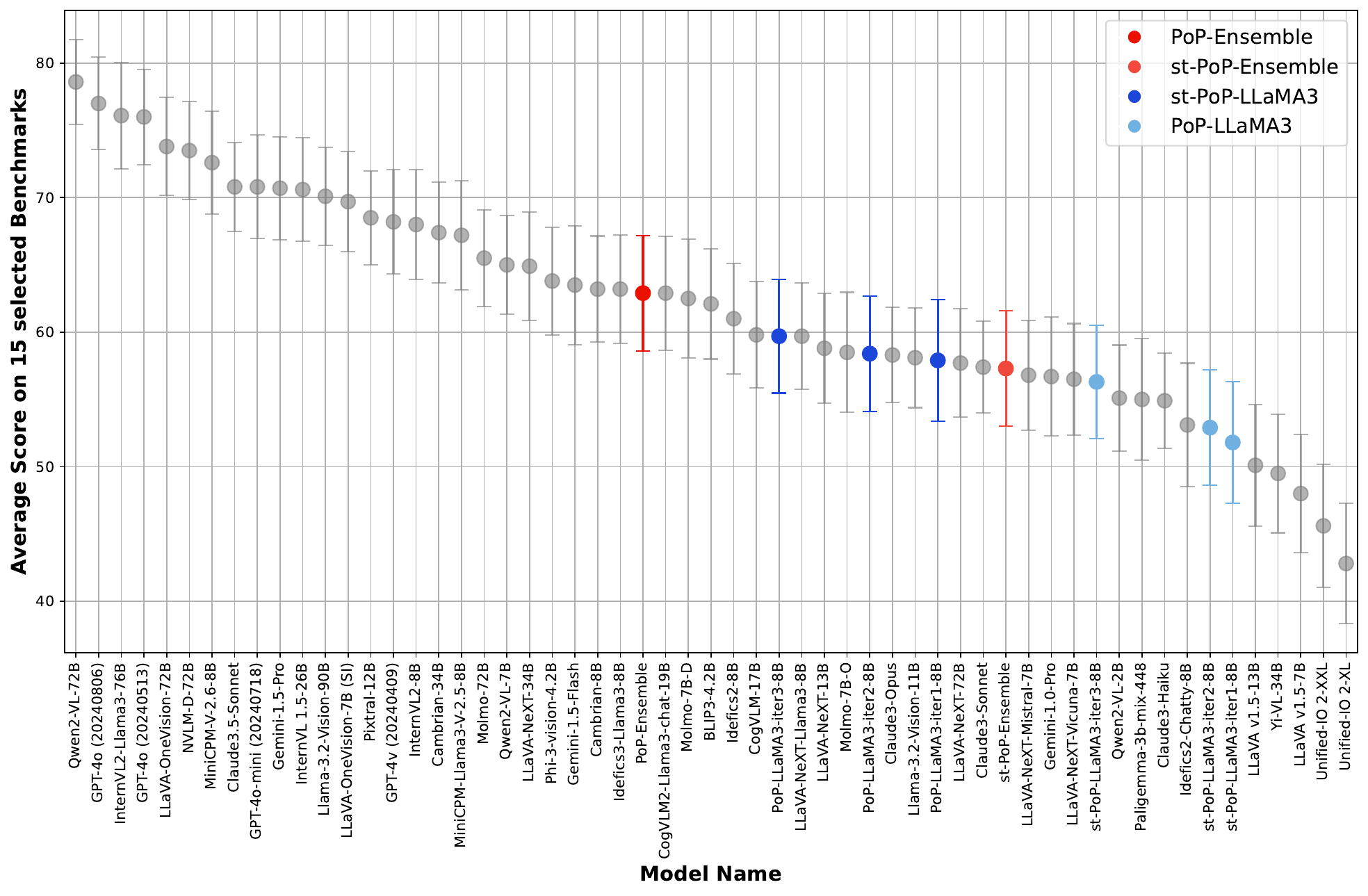}
    \caption{\textbf{Evaluation results of our approach on 15 selected benchmarks in the OpenVLM Leaderboard.}  The figure displays  49 selected LVLMs (until 2024.10.30) in descending order of average score. When calculating the average score, the scores of each benchmark are normalized to the range of 0 to 100.}
    \label{fig:comparisions}
\end{figure*}

\subsection{Extra Results on Learning and Ability from Scratch}
In addition to the ablation study presented in the main manuscript, where we evaluated the ability of the Panel-of-Peers (PoP) framework to teach a model OCR capabilities, we expanded the analysis to include the performance of the \emph{OCR-Dumb} model across other benchmark categories. Figure~\ref{fig:learn_ocr_full} provides a comprehensive view of the model's iterative performance improvement across five categories: \textit{Chart\&OCR}, \textit{General Knowledge}, \textit{Math and Science}, \textit{Hallucination}, and \textit{Vision-Centric} tasks.

The experiment began with an OCR-Dumb model trained with varying proportions of OCR knowledge (0\%, 25\%, 50\%, 75\%, and 100\%). Interestingly, the results demonstrate that as OCR knowledge increases, the model's performance steadily improves not only in OCR-related tasks but also in other categories. Notable observations include:
\begin{itemize}
    \item \textbf{Chart and OCR:} Performance rises sharply with increased OCR knowledge, validating the importance of reading capabilities for interpreting structured visual data.
    \item \textbf{General Knowledge:} Gains in this category suggest that improved text recognition contributes to better multimodal understanding and reasoning.
    \item \textbf{Math and Science:} Enhanced OCR capabilities positively impact tasks involving numerical and scientific reasoning, where understanding text is critical.
    \item \textbf{Hallucination:} Improvements here indicate that OCR knowledge helps reduce misalignments and inconsistencies in model outputs at the beginning. However, this improvement plateaus if the model starts with more OCR knowledge.
    \item \textbf{Vision-Centric:} Even tasks not directly reliant on OCR knowledge show gradual improvement, though to a lesser extent, with more OCR knowledge. This emphasizes the holistic impact of PoP training.
\end{itemize}

These results show the applicability of Peer-to-Peer Learning, demonstrating its ability to transfer knowledge, including OCR, while simultaneously increasing performance in various multimodal tasks. This highlights the effectiveness of PoP as a self-improvement mechanism, enabling models to iteratively learn new capabilities and address their initial weaknesses.

\definecolor{light-gray}{gray}{0.5}
\definecolor{light-green}{HTML}{D5F5E3}
\definecolor{light-blue}{HTML}{D6EAF8}

\begin{table*}[t!]
  \centering
  \small
  \setlength{\tabcolsep}{4pt}
  \renewcommand{\arraystretch}{1.2}
  \begin{tabular}{%
      ll                               % Capability, Benchmark
      ccc@{\hspace{6pt}}c@{\hspace{6pt}}% Iter-0  + spacer-1
      ccc@{\hspace{6pt}}c@{\hspace{6pt}}% Iter-1  + spacer-2
      ccc@{\hspace{6pt}}c@{\hspace{6pt}}% Iter-2  + spacer-3
      ccc}                             % Iter-3
  \toprule
  \multirow{2}{*}{\textbf{Capability}} &
  \multirow{2}{*}{\textbf{Benchmark}} &
  \multicolumn{3}{c}{\textbf{Iteration 0}} & \padcol &
  \multicolumn{3}{c}{\textbf{Iteration 1}} & \padcol &
  \multicolumn{3}{c}{\textbf{Iteration 2}} & \padcol &
  \multicolumn{3}{c}{\textbf{Iteration 3}}\\
  \cmidrule{3-5}\cmidrule{7-9}\cmidrule{11-13}\cmidrule{15-17}
        & & \MIS[3ex] & \VIC[3ex] & \LLA[3ex] & & % spacer-1
          \MIS[3ex] & \VIC[3ex] & \LLA[3ex] & & % spacer-2
          \MIS[3ex] & \VIC[3ex] & \LLA[3ex] & & % spacer-3
          \MIS[3ex] & \VIC[3ex] & \LLA[3ex] \\

% ─── sample row (showing the extra & for spacers) ─────────────
  \midrule
% General VQA
    \multirow{3}{*}{GeneralVQA} & MMBench~\cite{liu2023mmbench} 
    & 62.4 & 66.5 & 65.6 & & 65.3 & 65.5 & 68.7 && 65.7 & 66.1 & 69.9 && 67.0 & 67.4 & 71.3 \\
    & MM-Vet~\cite{yu2024mmvet}
    & 21.1 & 32.9 & 26.2 && 24.5 & 31.9 & 29.5 && 29.5 & 31.6 & 31.9 && 30.5 & 32.5 & 33.0 \\
    & SEED-Bench~\cite{li2024seed} 
    & 64.6 & 65.8 & 61.6 && 68.7 & 61.6 & 65.5 && 65.1 & 66.2 & 66.9 && 66.8 & 67.9 & 68.6 \\
    \midrule
% Knowledge
    \multirow{5}{*}{Knowledge} & $\dagger$AI2D~\cite{kembhavi2016diagram} 
    & 62.0 & 55.5 & 61.1 && 66.0 & 59.1 & 65.1 && 65.8 & 60.2 & 66.1 && 64.6 & 62.9 & 71.4 \\
    & MMMU~\cite{yue2024mmmu} 
    & 32.7 & 35.7 & 33.6 && 35.5 & 38.8 & 36.6 && 39.1 & 36.4 & 36.9 && 39.9 & 37.1 & 37.6 \\
    & MMStar~\cite{chen2024we}
    & 36.4 & 33.1 & 38.6 && 37.4 & 34.0 & 39.6 && 40.8 & 39.6 & 40.2 && 41.5 & 40.9 & 45.3\\
    & MathVista~\cite{lu2024mathvista} 
    & 30.3 & 25.6 & 30.3 && 33.1 & 31.2 & 33.1 && 34.9 & 33.8 & 35.5 && 35.0 & 34.9 & 37.7 \\
    & $\dagger$ScienceQA~\cite{lu2022learn} 
    & 58.0 & 66.8 & 71.2 && 62.4 & 67.1 & 73.1 && 66.1 & 71.9 & 75.4 && 68.0 & 74.0 & 77.6 \\
    \midrule
    
% Chart and OCR
    \multirow{4}{*}{Chart\&OCR} & $\dagger$ChartQA~\cite{masry2022chartqa} 
    & 39.6 & 31.9 & 40.4 && 42.4 & 42.7 & 43.3 && 46.3 & 45.1 & 45.7 && 48.4 & 47.1 & 47.8 \\
    & $\dagger$TextVQA~\cite{Singh_2019_CVPR} 
    & 44.9 & 45.5 & 44.9 && 48.4 & 49.0 & 48.4 && 50.2 & 50.3 & 49.3 && 52.2 & 52.3 & 51.2 \\
    & OCR-Bench~\cite{liu2023hidden} 
    & 33.6 & 31.8 & 33.9 && 34.7 & 33.8 & 35.0 && 39.5 & 38.7 & 38.3 && 41.3 & 41.6 & 44.5 \\
    & OCRVQA~\cite{mishra2019ocr} 
    & 59.7 & 60.6 & 57.7 && 62.7 & 63.6 & 60.6 && 60.9 & 62.4 & 61.7 && 61.4 & 62.9 & 62.2 \\
    \midrule
% Hallucination
    \multirow{2}{*}{Hallucination}  & POPE~\cite{li2023evaluating} 
    & 87.0 & 86.1 & 84.8 && 85.1 & 86.8 & 83.0 && 86.2 & 86.4 & 84.1 && 86.1 & 86.3 & 85.0 \\
    & HallusionBench~\cite{guan2023hallusionbench} 
    & 30.4 & 27.6 & 32.4 && 34.7 & 32.6 & 37.1 && 30.7 & 31.7 & 30.7 && 28.2 & 31.8 & 36.5 \\
    \midrule
% Vision centric
    Vision Centric & RWQA~\cite{realworldqa2024} 
    & 53.1 & 54.8 & 48.9 && 54.6 & 53.2 & 50.3 && 53.0 & 49.6 & 52.9 && 53.4 & 50.0 & 53.3 \\
    \midrule
    & \multicolumn{1}{r}{\textbf{Average}\mbox{  }} 
    & 47.7 & 48.0 & 48.7 && 50.4 & 50.1 & 51.2 && 51.6 & 51.3 & 52.4 && 51.2 & 51.6 & 53.7 \\
    \bottomrule
    \end{tabular}%
    % }
\caption{\textbf{Evaluation on 15 vision-language benchmarks.} We compare the performance of the single-try Panel-of-Peers (st-\pop). We have separated the benchmarks into five categories. Columns show three training iterations for \protect\MIS\ = \pop-Mistral, \protect\VIC\ = \pop-Vicuna, and \protect\LLA\ = \pop-LLaMA3. $\dagger$ indicates that the training set has been observed in our data mixture.}
\label{tab:iterations}
\vspace{-0.02in}
\end{table*}

\subsection{Extra Details on the Panel-of-Peers Ensemble as a Zero-Shot Evaluator}
We present more details on the experiments in Section 5.2. For models with more than 3B parameters, we included Phi-3-Vision~\cite{abdin2024phi}, BLIP3~\cite{xue2024xgen}, and Paligemma~\cite{beyer2024paligemma}. In the more than 7B range, we selected LLaVA-NeXT-Llama3, LLaVA-NeXT-Mistral, LLaVA-NeXT-Vicuna~\cite{liu2024llavanext}, and Idefics2~\cite{laurenccon2024matters}. For models exceeding 10B parameters, we picked CogVLM2-Chat~\cite{hong2024cogvlm2}, LLaVA-NeXT-Vicuna-13B~\cite{liu2024llavanext}, and Llama-3.2-Vision~\cite{llama3model}. For models with more than 30B parameters, we incorporated InternVL2-26B, InternVL 1.5-26B~\cite{chen2024far}, Cambrian-34B~\cite{tong2024cambrian}, and LLaVA-NeXT-Yi~\cite{liu2024llavanext}. Each panel performed response regeneration and evaluations. However, this is an evaluation-only method, enabling the creation of an ensemble using their consensus.

\subsection{Extra Results of Our Trained Models}
We present the specific scores of each of the members of the panel of peers, outlined in Table 2 of the main manuscript, where we presented the average scores of the whole panel.

 \fi

\end{document}